\documentclass[aos]{imsart}

\RequirePackage{amsthm,amsmath,amsfonts,amssymb}
\RequirePackage[numbers]{natbib}
\RequirePackage[colorlinks,citecolor=blue,urlcolor=blue]{hyperref}
\RequirePackage{graphicx}
\usepackage{amsmath}
\usepackage{tikz}
\usetikzlibrary{automata,arrows,positioning,calc}
\usepackage{verbatim}

\usepackage[linesnumbered,ruled,vlined]{algorithm2e}
\SetKwInput{KwInput}{Input}                
\SetKwInput{KwOutput}{Output}              

\startlocaldefs
\theoremstyle{plain}

\newtheorem{theorem}{Theorem}[section]
\newtheorem{lemma}[theorem]{Lemma}

\newtheorem{proposition}[theorem]{Proposition}

\newtheorem{remark}[theorem]{Remark}

\theoremstyle{remark}
\newtheorem{definition}[theorem]{Definition}
\newtheorem*{example}{Example}

\endlocaldefs

\begin{document}

\begin{frontmatter}
\title{Universal Approximation and the Topological Neural Network}
\runtitle{Topological Universal Approximation}

\begin{aug}
\author[A]{\fnms{Michael}~\snm{ A. Kouritzin}\ead[label=e1]{michaelk@ualberta.ca}},
\author[B]{\fnms{Daniel}~\snm{Richard}\ead[label=e2]{djrichar@ualberta.ca}
}
\address[A]{Department of Mathematical and Statistical Sciences,
University of Alberta\printead[presep={,\ }]{e1}}

\address[B]{Statistics Canada\printead[presep={,\ }]{e2}}
\end{aug}

\begin{abstract}
A topological neural network (TNN), which takes data from a Tychonoff 
topological space instead of the usual finite dimensional space, 
is introduced.
As a consequence, a distributional neural network (DNN)
that takes Borel measures as data is also introduced.
Combined these new neural networks facilitate things like
recognizing long range dependence, heavy tails and other properties
in stochastic process paths or like acting on belief states
produced by particle filtering or hidden Markov model algorithms.
The veracity of the TNN and DNN are then established herein by a 
strong universal approximation theorem for Tychonoff spaces and its corollary
for spaces of measures.
These theorems show that neural networks can arbitrarily approximate 
uniformly continuous functions (with respect to the sup metric) associated 
with a unique uniformity. 
We also provide some discussion showing that neural networks on positive-finite measures are a generalization of the recent deep learning notion of deep sets.
\end{abstract}



\end{frontmatter}

\setcounter{equation}{0}

\section{Introduction}\label{Intro}

Neural networks are not just computer implementations of vector
maps any more.
Indeed, \cite{Khoussi} used neural networks to design a classifier for a finite set of probability distributions.
\cite{Ma} trained a probability filter that produced conditional probability distributions which were then sent to a neural network using moment generating functions.
\cite{kratsios} provided universal approximation results for functions on finite dimensional topological spaces; that is, if $\mathcal{F}$ is dense in $C(\mathbb{R}^n ; \mathbb{R}^m)$, then $\{ f \circ \phi : f \in \mathcal{F} \}$ is dense in $C(X ; \mathbb{R}^m)$ when $\phi \colon X \to \mathbb{R}^n$ is continuous and injective.
\cite{lenzi} used distributional neural networks to estimate parameters for max-stable processes.
\cite{papamakarios2019} does density estimation using neural networks. He then frames likelihood-free inference as a density estimation problem. 
In their examples, \cite{papamakarios2019} only considers the density with respect to a set of intuitively picked statistics. 
Their density estimation amounts to having a neural network output the parameters for a mixed Gaussian distribution. 
The inputs of the neural network are summary statistics from the generated data.
\cite{chen2020} attempts to learn sufficient statistics for a data set. They use a fully connected network and leave the finding of an optimal network structure to future research, which is addressed within.
\cite{zaheer} develops the notion of deep sets. 
The purpose of this paper is to provide mathematical foundations to support
these works and future works where the input can even come from a fairly
general topological space.

Much of machine learning is about training networks to make good predictions based on past data. 
Let $X$ be the sets of possible predictor values with target values in 
$\mathbb{R}$; $p(x,y)$ be the population probability distribution over 
$X \times \mathbb{R}$; and $D = \{(x_i, y_i)\}_{i=1}^n \subset X \times \mathbb{R}$ be independent data samples from $p(x,y)$. 
The goal is to find a "good" predictor function $f \colon X \to \mathbb{R}$ such that $f(x_i)$ is "close" to $y_i$ for each $i = 1, \ldots, n$ from a collection $\mathcal{M}$ of possible predictor functions. 
Suppose $r \colon \mathbb{R} \times \mathbb{R} \to [0, \infty)$ is the selected error function, then the mean error for a predictor function across $p(x,y)$ is defined as
\begin{align}
    \widetilde{A}(f) = \int_{X \times \mathbb{R}} r( f(x), y) \, dp(x,y)
\end{align}
for which the best predictor function $g$ with respect to $p(x,y)$ is defined as
\begin{align}
    \widetilde{g} = \min_{f \in \mathcal{M}} \{ \widetilde{A}(f) \}, \label{theoretical_best_predictor}
\end{align}
assuming such $\widetilde{g}$ exists and is unique.
But, $p(x,y)$ is unknown so the empirical distribution is used in place of the population distribution and the average error for a predictor function across $D$ becomes
\begin{align}
    A_n(f) = \frac{1}{n} \sum_{i=1}^n r( f(x_i), y_i),
\end{align}
and the best predictor function $g$ with respect to D is given as
\begin{align}
   \widehat{g}_n = \min_{f \in \mathcal{M}} \{ A_n(f) \}. \label{theoretical_data_best_predictor}
\end{align}
One chooses $\mathcal{M}_{\Theta} $ as parameterized computer-workable functions 
$f_{\theta}$, 
where $\theta\in \mathbb{R}^d$ is a collection of real parameters controlling the behaviour of the predictor function so $\mathcal{M}_{\Theta} = \{f_{\theta} : \theta \in \Theta \}$. 
Now, the search for the best prediction function becomes a search for:
\begin{align}
    \widehat{\theta} = \min_{\theta \in \Theta} \{ A_n(f_{\theta})   \}. \label{practical_best_predictor}
\end{align}
If $\mathcal{M}_{\Theta}$ is uniform dense in $\mathcal{M}$, then one can be assured that 
$f_{\theta}$ can be made close enough to $\widehat{g}_n$ for some setting of $\theta$. 
\begin{definition}[Uniform Dense] \label{def:uniform_dense}
Let $\mathcal{F}$ and $\mathcal{G}$ be collections of real valued functions with common domain $X$. We say $\mathcal{F}$ is \textit{uniform dense} in $\mathcal{G}$ if and only if for each $g \in \mathcal{G}$ and $\epsilon > 0$, there exists an $f \in \mathcal{F}$ such that 
\begin{align}
    \sup \left\{ |f(x) - g(x)| : x \in X \right\} < \epsilon.
\end{align}
In addition, if $\mathcal{F} \subset \mathcal{G}$, then $\mathcal{F}$ is said to be a \textit{uniform dense subset of} $\mathcal{G}$.
\end{definition}
Often, one wishes for $\mathcal{M}_{\Theta}$ to be uniform dense within a collection of continuous functions of interest; in which case, $\mathcal{M}_{\Theta}$ is said to have the \emph{universal approximation property}. 

\cite{cybenko} showed that neural networks have the universal approximation property. In particular, he showed that functions of following form:
\begin{align}
     x \mapsto \sum_{j=1}^n \beta_j \sigma(a'_j x - \theta_j) && a_j \in \mathbb{R}^k; \beta_j, \theta_j \in \mathbb{R},
\end{align}
where $'$ denotes transpose and $\sigma$ is an $\mathbb R$-valued function\footnote{Additionally, $\sigma$ must be discriminatory, which is to say that $\int_{[0,1]^k} \sigma(a'_j x - \theta_j) \, d\mu = 0$ for each $a_j \in \mathbb{R}^k$ and $\theta_j \in \mathbb{R}$ implies $\mu = 0$.}, are uniform dense in the continuous functions defined on $[0,1]^k$.
Now, let $C(X)$ be the collection of real valued continuous functions on $X$, $f \vert_A$ be the restriction of $f$ to the subset $A \subset X$ and define:
\begin{definition}[Uniform Dense on Compacts] \label{def:uniform_dense_compacts}
Let $X$ be a topological space. Then, $\mathcal{F} \subset C(X)$ is said to be \textit{uniform dense on compacts of $X$} if for each compact $K \subset X$, $\{ f\vert _{K} : f \in \mathcal{F} \}$ is uniform dense in $C(K)$.
\end{definition}
\cite[Theorem 2]{hornik} extended the work of \cite{cybenko} to the compact subsets of $\mathbb{R}^k$:
\begin{theorem}(Hornik) \label{thm:hornik}
If $\sigma \colon \mathbb{R} \to \mathbb{R}$ is continuous, bounded and non-constant, then the following functions
\begin{align}
    \bigcup_{n \in \mathbb{N}}\mathcal F_n;\quad\text{where } \mathcal F_n=\left\{ x \mapsto \sum_{j=1}^n \beta_j \sigma(a'_j x - \theta_j) : a_j \in \mathbb{R}^k; \beta_j, \theta_j \in \mathbb{R} \right\}
\end{align}
are uniform dense on compacts of $\mathbb{R}^k$.
\end{theorem}
Theorem \ref{thm:hornik} has more recently been extended to continuous, non-polynomial $\sigma$.
Indeed, there are many collections of functions that are uniform dense of the compacts\footnote{This includes deep neural networks (see \cite{kidger} Theorem 3.2).} of $\mathbb{R}^k$. 
Their mere existence is importance to us.

The main goal of this paper is to motivate, develop and apply neural
networks with topological space inputs.
Let $\mathcal{R}_0(A)$ be the collection of all non-empty finite subsets of $A$
and  
$C(X;Y)$ be the continuous functions from topological space $X$ to topological space $Y$. 
$C(X)$ is then used when $Y = \mathbb{R}$ with standard topology and 
$C_B(X;Y)$, $C_B(X)$ are the bounded functions in $C(X;Y)$ and $C(X)$. 
\begin{definition}[Topological Neural Network]
Suppose $X$ is a topological space; $\mathcal{M} \subset C_B(X)$; and $\mathcal{F}_n \subset C(\mathbb{R}^n)$ for each $n \in \mathbb{N}$. 
Let $\mathfrak{N}( \mathcal{M}, \{ \mathcal{F}_n \}_{n=1}^{\infty})$ denote the functions:
\begin{align}
 \bigcup_{n =1}^{\infty}     \big\{ p \mapsto f (g_1(p), \, \ldots \, , g_n(p)) :  f \in \mathcal{F}_n; \; \{g_i\}_{i=1}^n \in \mathcal{R}_0(\mathcal{M}) \big\}. \label{topological_neural_network}
\end{align}
We call $\mathfrak{N}( \mathcal{M}, \{ \mathcal{F}_n \}_{n=1}^{\infty})$ the \textit{topological neural networks} generated by 
\textit{test functions} $\mathcal{M}$ and \textit{output networks} $\{ \mathcal{F}_n \}_{n=1}^{\infty}$.
\end{definition}

Next, we use topological neural networks to build distributional neural 
networks. 

\begin{definition}[Distributional Neural Network]\label{Def:DNN}
Suppose $E$ is a topological space;  $ g_{0} \in C_B((0, \infty)) $; {$\displaystyle \mathcal F_{n} \subset C(\mathbb R^{n}) $} for each $n \in \mathbb{N}$; and $\mathcal{M} \subset C_B(E)$. 
Let $\mathfrak{D}_{g_0}(\mathcal{\mathcal{M}}, \{ \mathcal{F}_n \}_{n=1}^{\infty})$ denote the mappings: 
\begin{align}
    \mu \mapsto f\left(g_{0}\left(\mu \left(E\right)\right),   \int_E  g_2\, \frac{d\mu}{\mu(E)}, \ldots , \int_E  g_n\, \frac{d\mu}{\mu(E)} \right), \label{distributional_neural_networks}
\end{align}
where $f \in \mathcal F_{n}$; $g_{2},...,g_{n} \in\mathcal  M$ ; and $n \in \mathbb{N}$.
We call $\mathfrak{D}_{g_0}( \mathcal{M},\{ \mathcal{F}_n \}_{n=1}^{\infty})$ the \textit{distributional neural networks} generated by test functions $\mathcal{M}$ and output networks $\{ \mathcal{F}_n \}_{n=1}^{\infty}$.
\end{definition}
\begin{remark}\label{FnExplain}
$\mathcal F_n$ can be any continuous functions in our definitions of the topological and distributional neural network.
However, since these networks will be implemented on a computer it is
usually the case that they are taken to be of the form given in
Theorem \ref{thm:hornik}.
\end{remark}

To motivate the infinite-dimensional setting, we consider the hidden Markov model:
\begin{align}
    X_0 = x_0 \sim p(x_0),\quad  
    X_i | X_{i-1} &= x_{i} \sim p(x_i \mid x_{i-1})   \\
    Y_i | X_{i} &= y_{i} \sim p(y_i \mid x_{i})
\end{align}
for $i = 1, \ldots, n$ and $x_i, y_i \in \mathbb{R}$. 
The $X_i$ are hidden (non-observed) random variables, while the $Y_i$ are observed. 
The goal is to compute the conditional distribution for the hidden variables given the observations $p(x_i \mid y_i, i < n)$. 
Particle filtering (see e.g. \cite{DMKoMi}, \cite{Ko17a}) is a common technique for computing the distribution by representing it as $m \in \mathbb{N}$ number of weighted particles. 
For $j=1,\ldots,m$ let $x^j_n \in \mathbb{R}$ and $L^j_n >0$ be the particle value and likelihood (given $y_i$ for $i < n$), respectively, for the $j$-th particle at time $n$. 
Then, the unnormalized measure based on the particles, given as
\begin{align}
    \mu(A) = \sum_{j=1}^m L^j_n \delta_{x_n^j}(A), \label{empirical_measure}
\end{align}
for measurable $A \subset \mathbb{R}$, is used to represent the target distribution as 
\begin{align}
    p(x_i \in A \mid y_i, i < n) \approx  \frac{\mu(A)}{\mu(\mathbb{R})}.
\end{align}
Both $\mu$ and $p(x_n \in A \mid y_i, i < n)$ provide information about the hidden variable 
$x_n$ given the past $y_i$'s. 
To utilize this information, we may want to learn a neural network decision function with a positive-finite measure as an input.
But, what class of predictor functions is appropriate for  
approximating continuous functions of positive-finite measures? 
\cite{Ma} (section 3.4) showed empirical success by evaluating the moment generating 
function for the input measure $\mu$ at various points, then passing those points into a neural network. 
We can compute the moment generating function for $\mu$ evaluated at $v \in \mathbb{R}$ as below: 
\begin{align}
M_{\mu}(v) = \int e^{v \cdot z} \, d\mu(z) 
= \sum_{i=1}^m L_n^j \cdot e^{v \cdot x_n^j}.
\end{align}
The resulting decision functions look like
\begin{align}
    \mu \mapsto f \left(M_{\mu}(v_1), \ldots , M_{\mu}(v_k)  \right),
\end{align}
where $f \in \mathcal{A}_k$ is a neural network on $\mathbb{R}^k$. 
In \cite{Ma}, the $v_1, \ldots, v_k$ are implemented as parameters, so the points in which they evaluate the moment generating function are learned.
It is one goal of this document to provide some theoretical justification for this practise by providing a universal approximation theorem for positive-finite measures. 

To motivate the infinite-dimensional setting further, we consider the path of
a stochastic process $\{X_t,\ t\in[0,T]\}$.
The path is cadlag and it is desired to estimate the heavy-tail and long-range dependence
parameters $\alpha\in(0,2]$ and $\sigma\in(0,1]$.
Because this path may have both heavy tails and long range dependence the standard tests
do not apply.
(The existence of heavy tails violates the known tests for long range dependence and
vice versa.)
However, the paths are in $E=D_{\mathbb R^k}[0,T]$, the cadlag functions endowed with the Skorohod 
$J_1$ topology, which is (separable, metrizable and) Tychonoff.
Hence, we will be able to apply the results developed herein.
In particular, we can take $\mathcal M$ to 
separate points (SP) and strongly separate points (SSP) on $E$.
For example, we can take 
$\mathcal M=\{\rho(\cdot,y_m):\{y_m\}_{m=1}^\infty\ \text{ dense in }E\}$,
where $\rho$ is a bounded metric consistent with the Skorohod topology.
Since $\mathcal M$ SP and SSP it can be used to construct a homeomorphism
between $E$, the Skorohod space here, and $\mathbb R^\infty$.
Also, $\mathcal M$ plays the role the moment generating functions did in
the particle filtering example.
Indeed, as will be proven herein, we can still use functions of the form 
$f(g_1,...,g_k)=f(\rho(\cdot,y_1),...,\rho(\cdot,y_k))$, where
$f$ is a normal neural network, as prediction functions, i.e. as neural
networks on Skorohod space $E$.
Indeed, we can use the $\{y_m\}$ as extra parameters and learn them as
we did with the moment generating function evaluation point $\{v_i\}$.  
We would train this \emph{topological neural network} on a large
selection of paths with known amounts of long range dependence and 
heavy tails and then apply it to unknown ones.
At this point, this application is purely hypothetical as details would
undoubtably be difficult.
However, the present work lays the mathematical foundation for this
and many other important future applications.

Deep sets, another application of our work, refer to neural network like 
functions whose inputs (or outputs) are sets. 
Some possible applications of deep sets include:
\begin{itemize}
    \item online shopping where a customer may purchase multiple items in a single online order,
    \item sports analytics where the goal is to understand the effectiveness of different lineup combinations of players in team sports, and
    \item a computer player for card games where players are dealt a hand of cards.
\end{itemize}

Our contributions are not just introducing
topological and distributional neural networks and proving universality
class results for them.
Our results are uniform approximations, with respect to a uniformity,
over a whole space and as such they provide new results even for
standard neural networks on $\mathbb R^d$.
When constructed according to the results herein neural networks on
$\mathbb R^d$ enjoy a universality class result on all of $\mathbb R^d$
not just compact subsets. 

The layout of this paper is as follows.
In the next section, the common examples of the sample functions
$\{g_i\}$ are given and it is explained how their parameters can
be learned in practice.
Section 3 contains our basic notation.
Section 4 contains our mathematical background and proofs
divided into subsections. 

\subsection{Notation List}

The following is a collection of the notation used within this paper,
most of which will be defined within. 

\begin{description}
\item[{$\setminus$}]  Set difference. $A\setminus B = \{x \in A: x \not \in B\}$.
\item[$f \vert_A$, $\mathcal{D}\vert_A$]  Restriction of $f$ to domain $A$. $\mathcal{D}\vert_A = \{f \vert_A : f \in \mathcal{D}\}$. 
\item[$x\vee y$, $x\wedge y$]  Binary $\max$ and $\min$ operators; equivalent to $\max\{x,y\}$ and $\min\{x,y\}$.
\item[$B(X,Y)$]  Collection of bounded functions from $X$ to $Y$. 
\item[$Y^I$] Cartesian Products $\prod_{i \in I} Y$. 
$Y^n$ is equivalent to $\prod_{i = 1}^n Y$ for $n \in \mathbb{N} \cup \{\infty\}$.
\item[$\pi_{I_0}$, $\pi_i$] Projection functions on Cartesian products. 
\item[$\bigotimes \mathcal{D}$] Function simultaneously evaluating all functions in $\mathcal{D}$ into the Cartesian product. 
\item[$\mathcal{O}(X)$, $\mathcal{C}(X)$] Open and closed sets on topological space $X$. 
\item[$\mathcal{O}_X(A)$]  Subspace topology induced on $A$ by $X$. 
\item[$\mathcal{B}_{\mathcal{D}}(A)$, $\mathcal{B}_{\rho}(A)$, $\mathcal{B}{[\mathcal{S}]}$] Topological basis induced on $A$ by functions $\mathcal{D}$ (See Proposition \ref{prop:basis_for_topology_generated_by_functions}), metric $\rho$ and subbasis $\mathcal{S}$ respecitvely. 
\item[$\mathfrak{Q}(X, \mathcal{T})$] The sequential topology on $X$ generated from the topology $\mathcal{T}$. See Definition \ref{def:gen_seq_top}. 
\item[$\mathfrak{cl}{[A]}$, $\overline{A}$] Topological closure of $A \subset X$ for some topological space $X$. 
\item[$\Delta(X)$] Diagonal of $X$. Defined as $\{(x,x) : x \in X \}$. 
\item[$A^{-1}$] Inverse relation of $A$. Defined as $\{(y,x) : (x,y) \in A\}$. 
\item[$A \circ B$] Composition of relations $A$ and $B$. Defined as  $\{(x,y) : \text{for some } z \in X, (x,z) \in A \text{ and } (z,y) \in B \}$. 
\item[$\mathcal{U}(X)$] The uniformity on the uniform space $X$. 
\item[$\mathcal{U}_X(A)$] The relative uniformity on $A$ induced by $X$ making $A$ a uniform subspace of $X$. 
\item[$\mathcal{U}_d(X)$] The metric uniformity $X$ generated by the metric $d$. 
\item[$C_U(X;Y)$, $C_U(X)$] Uniformly continuous functions from uniform space $X$ to uniform space $Y$. $C_U(X)$ implies $Y = \mathbb{R}$ with standard uniformity. 
\item[$\mathfrak{S}_{\mathcal{M}}$] Uniformity associated with the collection of functions $\mathcal{M}$. 
\item[$\mathfrak{B}(E)$] Borel sets of topological space $E$. Equivalent to $\sigma(\mathcal{O}(E))$. 
\item[$M(X;Y)$, $M(X)$] Measurable functions on measurable space $X$ to measurable space $Y$. $M(X)$ implies $Y = \mathbb{R}$ with $\sigma$-algebra $\mathfrak{B}(\mathbb{R})$. 
\item[$M_B(X;Y)$, $M_B(X)$] Bounded functions in $M(X;Y)$ and $M(X)$. 
\item[$\Rightarrow$] Weak convergence of positive-finite measures. See Definition \ref{def:weak_convergence}. 
\item[$\mathcal{P}(E)$, $\mathcal{M}^+(E)$] Collection of probability, positive-finite measures on measurable space $E$. 
\item[$f^*$ ] The mapping $f^*(\mu) \mapsto \int_E f \, d\mu$. 
$\mathcal{D}^*$ is defined as $\{f^* : f \in \mathcal{D} \}$. 
\item[$\mathcal{T}^{W}$] The weak topology of positive-finite measures. See Definition \ref{def:weak_topology_measures}. 
\item[$\mathcal{T}^{WC}$] The topology of weak convergence of positive-finite measures. See Definition \ref{def:topology_weak_convergence}. 
\item[$\mathfrak{W}_{g_0}{[\mathcal{M}]}$] Collection of functions on positive-finite measures generated by the collection of functions $\mathcal{M} \cup \{g_0\}$. See Proposition \ref{thm:blount_ssp_prob_measures}. 
\end{description}

\section{Motivation, Main Results and Examples}\label{Examples}

Suppose $E$ is a Tychonoff space and we wish to learn a function 
$\lambda \in C(E)$. 
Typically, one chooses a parameterized function $\lambda_{\Omega}$, where $\Omega \in \mathbb{R}^N$ represents $N$ parameters, and is led by data to an optimal $\widetilde{\Omega}$ such that $\lambda_{\widetilde{\Omega}}$ is \textit{closest} to $\lambda$.
We will use a functional approach to extend neural networks to larger and non-compact spaces based upon the following definitions:

\begin{definition}[Separation of Points]
Let $\mathcal{M}$ be a class of functions mapping $A$ to $B$. If for every $x,y \in A$ with $x \neq y$ there exists $f \in \mathcal{M}$ such that $f(x) \neq f(y)$, then $\mathcal{M}$ is said to \textit{separate points} (s.p.).
\end{definition}

\begin{definition}[Strong Separation of Points]
Let $(X, \mathcal{T})$ be a topological space and $\mathcal{M} \subset \mathbb{R}^X$ be a collection of $\mathbb R$-valued mappings. Then $\mathcal{M}$ \textit{strongly separates points} (s.s.p.) if, for every $x \in X$ and neighborhood $O_x$ of $x$, there is a finite collection $\{g_1, \ldots, g_k \} \in \mathcal{R}_0(\mathcal{M})$ such that 
\begin{align}
    \inf_{y \not \in O_x} \max_{1 \leq i \leq k} |g_i(y) - g_i(x)| > 0. \label{ssp_criteria}
\end{align}
\end{definition}

We first give some general examples of function classes that s.p. and s.s.p.
\begin{example}
When $X$ is a Tychonoff space, Proposition \ref{lem:tychonoff_equiv_sep_and_ssp} (to follow) tells us $C_B(X)$ s.s.p. and s.p. on $X$. So any uniformly dense subset of $C_B(X)$ s.p. and s.s.p.$\quad \square$
\end{example}

The next example is interesting because each function is non-zero in a bounded region, so each function can be "turned off" depending on the input. 
\begin{example}[From \cite{blount}]
Suppose $(X, d)$ is a metric space. Then, the following collection of functions are continuous, s.p., and s.s.p. on $X$:
\begin{align}
    \{ g_{q,k}(p) \mapsto (1 - k d(p,q)) \vee 0 : q \in X, k \in \mathbb{N}\}.
\quad \square
\end{align}
\end{example}

Finally, we conclude with Hibert spaces.
By Proposition \ref{prop:ssp_homeomorphism_determine_point_convergence} (to follow)
to test a class of functions $\mathcal M$ s.s.p. on a first countable Hausdorff
space we need only show that $g(x_n)\rightarrow g(x)$ for all $g\in \mathcal M$
implies $x_n\rightarrow x$.

\begin{example}
Suppose $(X,\langle \cdot , \cdot \rangle)$ is a Hilbert space with a countable complete orthonormal basis $\mathcal{B}$ but endowed with the weak topology. 
Define $g_e(p) \mapsto \langle p , e \rangle$ and let $\mathcal{M} = \{g_e  : e \in \mathcal B \}\subset C(X)$. If $p \neq q$, then 
\begin{align}
    p - q &= \sum_{e \in \mathcal B} \langle p - q ,  e \rangle e \neq 0,
\end{align}
so there is some $e \in \mathcal{B}$ such that $\langle p - q ,  e \rangle = g_e(p) - g_e(q) \neq 0$, implying $\mathcal{M}$ s.p. on $X$.  Now suppose $(x_n)$ is a sequence in $X$ such that $g_e(x_n) \to g_e(p)$ for each $e \in \mathcal{B}$. That is, we have $\langle x_n , e \rangle \to \langle p , e \rangle$
which implies $x_n\rightarrow x$ with the weak topology
and by the comment just prior to the example we conclude that $\mathcal{M}$ s.s.p. on $X\quad\square$.
\end{example}

The s.p. property implies a bijection and
s.s.p. converts this bijection to an imbedding of $E$ into $\mathbb R^{\mathcal M}$, which allows us to compactify the input space.

\begin{definition}[Compactification, Equivalent, Unique]
Suppose $X$ is a topological subspace of a compact space $S$. Then, $S$ is called a \textit{compactification} of $X$ if $\overline{X} = S$. 
If $S$ and $T$ are compactifications of $X$, then we say they are \textit{equivalent up to homeomorphism} if there exists a homeomorphism $h \colon S \to T$ such that $h(p) = p$ for each $p \in X$. If every compactification of $X$ is equivalent, then $S$ is \textit{unique up to homeomorphism}. 
\end{definition}

\begin{proposition}
\label{lem:spp_equiv_compact}
Let $X$ be a topological space and $\mathcal{M} \subset \mathbb{R}^{X}$ be a collection of bounded functions. Then, the following statements are equivalent:
\begin{enumerate}
    \item $\mathcal{M} \subset C_B(X)$ separates and strongly separates points on $X$.
    \item X admits a unique compactification $S$ up to homeomorphism such that $\bigotimes \mathcal{M}$ extends to a homeomorphism between $S$ and the closure of $\bigotimes \mathcal{M}(X)$ in $\mathbb{R}^{\mathcal{M}}$.
    \item $\bigotimes \mathcal{M}$ is an imbedding of $X$ in $\mathbb{R}^{\mathcal{M}}$.
\end{enumerate}
\end{proposition}
\begin{proof}
We provide the proof from \cite{dong_kouritzin} Lemma 9.3.4.

($1 \to 2$)  \cite{kouritzin2016tightness} Theorem 6 (1 - 3) shows that there exists a compact $S$ and homeomorphism $h \colon S \to \mathfrak{cl}[\bigotimes \mathcal{M}(X)]$ such that $h \vert_X = \bigotimes \mathcal{M}$. We show $S$ is unique up to homeomorphism. Suppose $T$ is another compactification of $X$ such that $b \colon T \to \mathfrak{cl}[\bigotimes \mathcal{M}(X)]$ such that $b \vert_X = \bigotimes \mathcal{M}$. Then $b^{-1} \circ h \colon S \to T$ is a homeomorphism such that $b^{-1} \circ h (p) = p$ for each $p \in X$, which implies $S$ and $T$ are equivalent.

($2 \to 3$) Is proven directly as $X$ is a subspace of $S$.

($3 \to 1$) Since the product of Tychonoff spaces is Tychonoff $\bigotimes \mathcal{M}$ is injective and $\mathcal{O}(X) = \mathcal{O}_{\mathcal{M}}(X)$. Given $p \neq q \in X$, $\bigotimes \mathcal{M}(p) \neq \bigotimes \mathcal{M}(q)$; hence, $\pi_f \circ \bigotimes \mathcal{M}(p) \neq \pi_f \circ \bigotimes \mathcal{M}(q)$ for some $f \in \mathcal{M}$ and $\pi_f \circ \bigotimes \mathcal{M} = f$, so $\mathcal{M}$ s.p. on $X$. Finally, $\mathcal{M}$ s.s.p. by Proposition \ref{prop:ssp_equiv_to_finer_topology} to follow.
\end{proof}

The compactification $S$ and the associated homeomorphism will be vital when proving our universal approximation results.
Interestingly, when there is a countable collection of functions that s.p. and 
s.s.p., there is a metric on the compactification $S$. 

\begin{proposition} \label{lem:countable_sep_spp_has_this_metric}
Let $X$ be a topological space; $N \in \mathbb{N} \cup \{\infty\}$; $\mathcal{M} = \{g_i\}_{i=1}^{N} \subset C(X)$  s.p. and s.s.p. on $X$; and let $\overline{h} \colon S \to \mathfrak{cl}[\bigotimes \mathcal{M}(X)]$ denote the extended homeomorphism mentioned in Proposition \ref{lem:spp_equiv_compact} (2). Then, $S$ is metrized by the following metric:
\begin{align}
    d(x,y) \mapsto \sum_{i=1}^{N} 2^{-i} \left( |\overline{g}_i(x) - \overline{g}_i(y)| \wedge 1 \right) && \forall x,y \in S \label{countable_metric}
\end{align}
where $\overline{g}_i \doteq \pi_i \circ h$ for each $i= 1 , \ldots, N$.
\end{proposition}
\begin{proof}
See \cite{kouritzin2016tightness} Theorem 6 (4).
\end{proof}

Our main result is the following Topological Neural Network, which is proved after our discussion on uniform spaces.
\begin{theorem}
Suppose $X$ is a topological space; $\mathcal{M} \subset C_B(X)$ separate and strongly separate points on $X$; and, for each $n \in \mathbb{N}$, $\mathcal{F}_n$ is uniform dense on compacts of $\mathbb{R}^n$. Then $\mathfrak{N}( \mathcal{M}, \{ \mathcal{F}_n \}_{n=1}^{\infty})$
is a uniform dense subset of $C_U(X,\mathfrak{S}_{\mathcal{M}})$. Additionally, if $\mathcal{M}$ is countable with cardinality $N \in \mathbb{N} \cup \{\infty\}$, then $\mathfrak{S}_{\mathcal{M}}$ is equivalent to the metric uniformity generated by the following metric:
\begin{align}
    d(x,y) \mapsto \sum_{i=1}^{N} 2^{-i} \left( |g_i(x) - g_i(y)| \wedge 1 \right) && \forall x,y \in X.
\end{align}
\end{theorem}

If the function to learn or approximate is $\lambda \in C_U(\mathcal{M}^+(E))$, 
then our first main Distributional Neural Network result, proved later, provides the parameterized $\{\lambda_{\Omega}:\Omega \in \mathbb{R}^N\}$:
\begin{theorem} \label{thm:DNN}
Suppose $N\in\mathbb N\cup\{\infty\}$; $E$ is a topological space;  {$\displaystyle g_{0} \in C_B((0,\infty)) $} s.p. and s.s.p. on {$\displaystyle (0, \infty) $}; {$\displaystyle \mathcal M = \{g_i \}_{i=2}^N \subset C_{B}\left(E\right) $} s.p., s.s.p., is countable and closed under multiplication; and {$\displaystyle \mathcal F_{n}\subset C(\mathbb R^{n}) $} is uniform dense on the compacts of {$\displaystyle \mathbb{R}^{n} $} for each {$\displaystyle n\in \mathbb{N} $}. 
Then $\mathfrak{D}_{g_0}(\mathcal{\mathcal{M}}, \{ \mathcal{F}_n \}_{n=1}^{\infty})$ is a uniform dense subset of {$\displaystyle C_U \big(\mathcal{M}^+(E),\mathfrak{S}_{\mathfrak{W}_{g_0}[\mathcal{M}]} \big) $}. Additionally, $\mathfrak{S}_{\mathfrak{W}_{g_0}[\mathcal{M}]}$ is equivalent to the metric uniformity generated by the metric:
\begin{align}
    d(\mu,\nu) \mapsto 
    \left(
    \begin{aligned}
    &|g_0(\mu(E)) - g_0(\nu(E))| \wedge 1 \\ + \sum_{i=2}^N 2^{-i}& \left( \left| g_i^*\left( \frac{\mu}{\mu(E)} \right) - g_i^*\left( \frac{\nu}{\nu(E)} \right) \right| \wedge 1 \right)
    \end{aligned}
    \right)
\end{align}
for each $ \mu,\nu \in \mathcal{M}^+(E).$
\end{theorem}

According to this theorem, Definition \ref{Def:DNN} and Remark \ref{FnExplain}, there are decisions when constructing our set $\{\lambda_{\Omega} : \Omega \in \mathbb{R}^N \}$. We need to choose $ n \in \mathbb{N}$, an activation function $\sigma$ and functions $(g_1, \ldots, g_n) \subset \mathcal{M}$.
This last decision is often difficult as $\mathcal{M}$ is countably infinite. 

\begin{example} \label{ex:measureNN1}
Suppose $E = [0,1]$, $x \in E$ and $g_{i+1}(x) = x^i$. $g_2$ s.s.p. and s.p. on $E$ as the identity function is a homeomorphism. 
Define $\mathcal{M} = \{g_{i+1} : i \in \mathbb{N}\}$. 
Therefore, $\mathcal{M} \subset C_B(E)$ is countable, closed under multiplication, s.p and s.s.p on $E$. Let $g_0(x) = \arctan(x)$, which is 1-1 and s.s.p on $\mathbb{R}$ (as it is a homeomorphism) and we use the result from Hornik in Theorem $\ref{thm:hornik}$ to select as our $\mathcal{F}_n$.
Putting this all together, we have that functions of the following form
\begin{align}
    \sum_{j=1}^n \beta_j \sigma \left(a_j'\left(\arctan\left(\mu \left([0,1]\right)\right),\int x^1\frac{d\mu }{\mu \left([0,1]\right)},...,\int x^{n-1}\frac{d\mu }{\mu \left([0,1]\right)}\right) - \theta_j \right),
\end{align}
where $n \in \mathbb{N}$, $a_j \in \mathbb{R}^{n}$, and $\beta_j, \theta_j \in \mathbb{R}$;
are uniformly dense in $C_{U}(\mathcal{M}^+([0,1]), \mathfrak{S}_{\mathfrak{W}_{g_0}[\mathcal{M}]})$ by Theorem \ref{thm:DNN}. $\ \ \square$
\end{example}
In this example, we just took the small powers $(x^1, x^2, \ldots, x^{n-1})$; however, in practice it can be beneficial to learn the $g$ functions like the network.
Specifically, we let our distributional neural network test functions be of the form
$\mathcal{M}=\{x^m:m\in\mathbb N\}$ and learn the best $(m_1,...,m_{n-1})$
along with the weights and biases of the output network. 

To generalize this example to higher dimensions, we suppose $D \in \mathbb{N}$, $E = [0,1]^D$ and
\begin{align}
    \mathcal{M} = \left\{ \prod_{i=1}^D x_i^{d_i} : d_i \in \mathbb{N} \cup \{0\} \right\}
\end{align}
which is closed under multiplication, s.p., and s.s.p. on $[0,1]^D$. 

\begin{example} \label{ex:measureNN2}
Suppose $E = [0,1]^D$, $V=(v_1,...,v_D)$ is a non-singular matrix and, 
inspired by the moment generating function and \cite{Ma},
$\mathcal M$ is the functions of the form 
\begin{align}
    x \mapsto e^{v'x}
\end{align}
where $v$ is a finite sum of the columns of $V$ and $'$ denotes transpose.
Since any column could be in the sum multiple times these functions are
closed under multiplication.
Since $V$ is non-singular these functions s.p. and s.s.p. 
Therefore, similar to the previous example, we can approximate the uniformly continuous functions on $\mathcal{M}^+([0,1]^D)$. $\ \ \square$
\end{example}

The distributional neural network is also related to the new and exciting area of
deep sets.
In the following discussion, we will only focus on deep sets where the input is a set. 

\begin{example}
\cite{zaheer} and \cite{wagstaff} study neural networks to approximate real-valued set functions on the finite subsets of ${[0,1]}$ through permutation invariance. 
A function $t \colon [0,1]^n \to \mathbb{R}$ is \textit{permutation invariant} if, for any permutation $p$ of $n$ elements, 
\begin{align}
    t(x_1, \ldots, x_n) = t(x_{p(1)}, \ldots, x_{p(n)}),
\end{align}
with the intuition being that the order of objects in a set is irrelevant. 
\cite[Theorem 7]{zaheer} identifies the permutation invariant functions in the following result:
\begin{proposition} \label{prop:perm_invariance_rep}
A function $t \colon [0,1]^n \to \mathbb{R}$ is permutation invariant if and only if it can be represented as
\begin{align}
    t(x_1, \ldots, x_n) = f \left( \sum_{i=1}^n g(x_i) \right), \label{perm_invariant_funcs_rep}
\end{align}
for some continuous functions $g \colon [0,1] \to \mathbb{R}^{n+1}$ and $f \colon \mathbb{R}^{n+1} \to \mathbb{R}$.
\end{proposition}
However, if:
$
    \mu(A) = \sum_{i=1}^n  I_{x_i}(A), \label{particle_measure}
$
for measurable $A \in \mathcal{B}([0,1])$, then
(\ref{perm_invariant_funcs_rep}) becomes
\begin{align}
    t(x_1, \ldots, x_n) = f \left( \sum_{i=1}^n g(x_i) \right)=f\left( \int_{[0,1]} g\, d\mu \right)=f\left(\mu([0,1]) \int_{[0,1]} g\, \frac{d\mu}{\mu([0,1])} \right)\, \label{perm_invariant_int_rep}
\end{align}
which is a function of a (unnormalized) distribution of the form described in 
the neural networks (\ref{distributional_neural_networks}). 
Further, the polynomials
are dense in $C[0,1]$ and the monomials s.p., s.s.p. and are closed under multiplication.
Hence, by linearity of integration there are continuous $f$ (different than above) and monomials $g_1,...,g_n$ such that 
\begin{align}
     t(x_1, \ldots, x_n) \approx f\left(g_{0}\left(\mu \left([0,1]\right)\right),   \int_{[0,1]}  g_1\, \frac{d\mu}{\mu([0,1])}, \ldots , \int_{[0,1]}  g_n\, \frac{d\mu}{\mu([0,1])} \right) \label{deep_neural_networks}
\end{align}
arbitrarily closely.
This same form is good for any permutation invariant function and the approximation network, being a function of only the measure $\mu$, is also permutation invariant.

Deep sets (with set inputs) can be handled by Theorem \ref{thm:DNN}.
In fact, it is sensible to think of deep sets as functions on spaces of positive-finite measures. 
Theorem 3.3 of  \cite{wagstaff} shows there exist set functions $t \colon 2^{[0,1]} \to \mathbb{R}$ that cannot be represented in the form of (\ref{perm_invariant_funcs_rep}), which is not surprising since there are uncountably infinite subsets of $[0,1]$. 
However, the theory developed herein can handle (measurable) infinite sets.  
Suppose $B \in \mathcal{B}([0,1])$ is a set with positive Lebesgue measure we would like to input to a neural network. We can represent it as a measure defined as
\begin{align}
    \mu_B(A) \mapsto \ell(A \cap B),
\end{align}
where $\ell$ is the Lebesgue measure, and results in the following functions
\begin{align}
    h & \left( g_0(\mu_B([0,1])),  \int_{[0,1]} f ( z ) \, d\mu_B(z) \right) 
\label{uncountable_funcs}
    =  h \left( g_0(\ell(B)),  \int_{B} f ( z ) \, d\ell(z) \right) .\quad\square
\end{align}
\end{example}

Sometimes it is not natural for our class $\mathcal M$ to be closed under multiplication. 
In these cases, we use our second main distribution neural network result, also proved later.

\begin{theorem} \label{thm:DNN1}
Suppose $E$ is a topological space;  $g_{0} \in C_B(E) $ s.p. and s.s.p. on $(0, \infty)$; $ \mathcal M  $ is countable, s.p. and s.s.p. on $E$; and $\mathcal F_{n}, \mathcal{H}_n \subset C(\mathbb R^{n}) $ are uniform dense on the compacts of $\mathbb{R}^{n}$ for each $n \in \mathbb{N}$.
Then
$\mathfrak{D}_{g_0}( \mathfrak{N} (\mathcal{M}, \{ \mathcal{F}_n \}_{n=1}^{\infty}), \{ \mathcal{H}_n \}_{n=1}^{\infty})$ is a uniform dense subset of {$\displaystyle C_U \big(\mathcal{M}^+(E),\mathfrak{S}_{\mathfrak{W}_{g_0}[C_U(E,\mathfrak{S}_{\mathcal{M}})]} \big) $}.
\end{theorem}

\begin{example} \label{ex:practical_example}

Suppose $D \in \mathbb{N}$, $E = [0,1]^D$, and $\mathcal{M} = \{\pi_1, \ldots, \pi_D\}$. Therefore, $\mathcal{M} \subset C_B(E)$ is countable, s.p and s.s.p on $E$. Let $g_0(x) = \arctan(x)$, which is bounded, continuous, s.p., and s.s.p on $\mathbb{R}$ and we represent functions by Theorem \ref{thm:DNN1} as
\begin{align}
    h \left( g_0(\mu(E)),  \int_E f ( g_2, \ldots, g_n ) \, d\mu \right), \label{multi_dimensional_g_representation}
\end{align}
where $f\in\mathcal{F}_n$ and $h\in\mathcal{H}_n$ can be the neural networks described by Hornik in Theorem \ref{thm:hornik}. However, we use the multidimensional version of Hornik for the $\mathcal{F}_n$. 
This leads to choose a particular $n_1, n_2, m \in \mathbb{N}$, and activation function $\sigma$. In order to construct our $\lambda_{\Omega}$, we first define the following function $\psi_{{A}, {B}, \Theta} \colon \mathcal{M}^+(E) \to \mathbb{R}^{1+m}$ as
\begin{align}
\psi_{{A}, {B}, \Theta}(\mu) \doteq
\begin{bmatrix}
\arctan\left(\mu \left([0,1]^D\right)\right) \\
\displaystyle \int_{[0,1]^D}{{B} \sigma \left( {A}' \left[\begin{smallmatrix}
  \pi_1(x) \\
  \vdots \\
  \pi_D(x)
\end{smallmatrix}\right] - \Theta \right)\frac{d\mu }{\mu \left([0,1]^D\right)}}
\end{bmatrix},
\end{align}
which results in the following set of functions $\{\lambda_{\Omega} : \Omega \in \mathbb{R}^N \}$ given as
\begin{align} \label{ex:practical_functions}
    \left\{ 
    \begin{aligned}
    \sum_{j=1}^{n_2} q_j \sigma \left(p_j' \, \psi_{{A}, {B}, \Theta}(\mu) - \phi_j \right):
    {A} \in \mathbb{R}^{D \times n_1}; \;\; {B} \in \mathbb{R}^{m \times n_1} \\ \;\; \Theta \in \mathbb{R}^{n_1} ; \;\; p_j \in \mathbb{R}^{1+m}; \;\; q_j, \phi_j \in \mathbb{R} 
    \end{aligned}
    \right\},
\end{align}
where $\sigma$ operates elementwise on vectors; that is, 
$\sigma(\big[\begin{smallmatrix}
  x_1 \\
  x_2 \\
  x_3
\end{smallmatrix}\big]) = 
\left[\begin{smallmatrix}
  \sigma(x_1) \\
  \sigma(x_2) \\
  \sigma(x_3)
\end{smallmatrix}\right]. 
$
All of the $A$, $B$, $ \Theta$, $\phi$, $p, q$ can be learnt.
$\ \square$
\end{example}

\section{Background on Spaces}\label{Notation}

\subsection{Topological Spaces}
We utilize topologies induced 
from collections of functions.

\begin{definition} \label{def:toplogy_define_by_functions}
Let $X$ be a set and $A \subset X$. For an index set $I$, let $Y_i$ be a topological space and $f_i \colon X \to Y_i$ for each $i \in I$. The \textit{topology induced by $\mathcal{D} \doteq \{f_i\}_{i \in I}$ on $A \subset X$}, denoted by $\mathcal{O}_{\mathcal{D}}(A)$, is the one generated from the subbasis
\begin{align}
    \mathcal{S}_{\mathcal{D}}(A) \doteq \left\{ f_i^{-1}(O) \cap A : O \in \mathcal{O}(Y_i),  i \in I \right\}. \label{subbasis_by_funcs}
\end{align}
\end{definition}
\begin{remark}
For any $O \in \mathcal{O}(Y_i)$ and $f_i \in \mathcal{D}$, we have $f_i^{-1}(O) \cap A \in \mathcal{S}_{\mathcal{D}}(A) \subset \mathcal{O}_{\mathcal{D}}(A)$. Therefore, $f\vert_A \in C((A, \mathcal{O}_{\mathcal{D}}(A)), Y_i)$. That is, for any collection of functions $\mathcal{D}$, we are able to generate a topology on $A$ such that $\mathcal{D}\vert_A$ are continuous. Further, $\mathcal{O}_{\mathcal{D}}(A)$ is the coarsest topology such that $\mathcal{D}\vert_A$ are continuous.
\end{remark}

\begin{proposition} \label{prop:subbasis_for_O_D}
Let $X$ be a set and $A \subset X$. For an index set $I$, let $Y_i$ be a topological space with subbasis $\mathcal{S}_i$ and $f_i \colon X \to Y_i$ for each $i \in I$. Set $\mathcal{D} \doteq \{f_i\}_{i \in I}$.
Then, 
\begin{align}
    \mathcal{S}_{\mathcal{D}}(A ; \{\mathcal{S}_i\}_{i \in I}) \doteq \left\{ f_i^{-1}(O) \cap A : O \in \mathcal{S}_i,  i \in I \right\}
    \label{subbasis_by_funcs_subbsises}
\end{align}
is a subbasis on $A \subset X$ that generates $\mathcal{O}_{\mathcal{D}}(A)$.
\end{proposition}

\begin{proof}
$\mathcal{S}_{\mathcal{D}}(A ;\{\mathcal{S}_i\}_{i \in I})$ is a subbasis for a topology on $A$ since 
\begin{align}
    \bigcup_{V \in \mathcal{S}_{\mathcal{D}}(A ;\{\mathcal{S}\}_{i \in I})} V &= \bigcup_{i \in I} f_i^{-1} \left( \bigcup_{O \in \mathcal{S}_i} O \right) \cap A  \\\nonumber
    &= \bigcup_{i \in I} f_i^{-1} \left( Y_i \right) \cap A \\\nonumber
    &= A.
\end{align}
Now, let $\mathcal{T}$ denote the topology generated from the subbasis $\mathcal{S}_{\mathcal{D}}(A ;\{\mathcal{S}_i\}_{i \in I})$ and $\mathcal{B}[\mathcal{S}_{\mathcal{D}}(A ;\{\mathcal{S}_i\}_{i \in I})]$ and $\mathcal{B}[\mathcal{S}_{\mathcal{D}}(A ; \{\mathcal{O}(Y_i)\}_{i \in I})]$ be the bases for $\mathcal{T}$ and $\mathcal{O}_{\mathcal{D}}(A)$, respectively. 

$\mathcal{B}[\mathcal{S}_{\mathcal{D}}(A ;\{\mathcal{S}_i\}_{i \in I})] \subset \mathcal{B}[\mathcal{S}_{\mathcal{D}}(A ;\{\mathcal{O}(Y_i)\}_{i \in I})]$ since $\mathcal{S}_i \subset \mathcal{O}(Y_i)$,  so $\mathcal{T} \subset \mathcal{O}_{\mathcal{D}}(A)$. 

Conversely, suppose $U \in \mathcal{B}[\mathcal{S}_{\mathcal{D}}(A ; \{\mathcal{O}(Y_i)\}_{i \in I})]$ and $x \in U$. Then, for some $I_0 \in \mathcal{R}_0 [I]$, $U$ has the form
\begin{align}
    U = \bigcap_{i \in I_0} f_i^{-1}(O_i) \cap A
\end{align}
where $O_i \in \mathcal{O}(Y_i)$ for each $i \in I_0$. For each $i \in I_0$, we have $f_i(x) \in O_i$; hence, there is a basis element $B_i \in \mathcal{B}[\mathcal{S}_i]$ such that $f_i(x) \in B_i \subset O_i$. But, $B_i = \bigcap_{V \in \mathcal{V}_i} V$ for some $\mathcal{V}_i \in \mathcal{R}_0[\mathcal{S}_i]$ so
\begin{align}
    x \in \bigcap_{i \in I_0} f_i^{-1}(B_i)   
   =\bigcap_{i \in I_0}& \bigcap_{V \in \mathcal{V}_i} f_i^{-1}\left( V \right) \cap A \\\nonumber
    &\in \mathcal{B}[\mathcal{S}_{\mathcal{D}}(A ; \{\mathcal{S}_i\}_{i \in I})],
\end{align}
so $\mathcal{T} \supset \mathcal{O}_{\mathcal{D}}(A)$, which completes the proof.
\end{proof}

In the case each $Y_i=\mathbb R$ with the standard topology, a subbasis for $\mathcal{O}_{\mathcal{D}}(A)$ is given as
\begin{align}
    \mathcal{S}_{\mathcal{D}}&(A ;\mathcal{B}_{\rho}(\mathbb{R})) \doteq \left\{ f^{-1}(O) \cap A : O \in \mathcal{B}_{\rho}(\mathbb{R}), f \in \mathcal{D} \right\} \label{subbasis_gen_by_funcs} \\\nonumber
    &= \bigg\{ \big\{x \in X : \left| a - f(x) \right| < \epsilon \big\} \cap A : a \in \mathbb{R}, \epsilon > 0, f \in \mathcal{D} \bigg\}
\end{align}
which then generates the basis $\mathcal{B}[\mathcal{S}_{\mathcal{D}}(A ;\mathcal{B}_{\rho}(\mathbb{R}))]$ given by the following sets
\begin{align}
    \bigg\{ \bigcap_{i=1}^n \big\{x \in X : \left| a_i - f_i(x) \right| < \epsilon_i \big\} \cap A : a_i \in \mathbb{R}, \epsilon_i > 0, f_i \in \mathcal{D}, n \in \mathbb{N} \bigg\}. \label{basis_gen_by_funcs}
\end{align}

\begin{proposition} \label{prop:basis_for_topology_generated_by_functions}
Suppose $X$ is a topological space, $A \subset X$, and $\mathcal{D} \subset \mathbb{R}^X$ where $\mathbb{R}$ is given the standard topology.
The following collection of sets
\begin{align}
    \bigg\{ \big\{x \in A : \max_{1 \leq i \leq n} \left| f_i(y) - f_i(x) \right| < \epsilon \big\}  : y \in A, \epsilon > 0, f_i \in \mathcal{D}, n \in \mathbb{N} \bigg\},
\end{align}
denoted as $\mathcal{B}_{\mathcal{D}}(A)$, is a basis for $\mathcal{O}_{\mathcal{D}}(A)$.
\end{proposition}
\begin{proof}
Notice the following
\begin{align}
    \big\{x \in A :  \max_{1 \leq i \leq n} & \left| f_i(y) - f_i(x) \right| < \epsilon \big\} 
    &\!\!\!= \bigcap_{i=1}^n \big\{x \in X : \left| f_i(y) - f_i(x) \right| < \epsilon \big\} \cap A.
\end{align}
Therefore, $\mathcal{B}_{\mathcal{D}}(A) \subset \mathcal{B}[\mathcal{S}_{\mathcal{D}}(A ;\mathcal{B}_{\rho}(\mathbb{R}))] \subset \mathcal{O}_{\mathcal{D}}(A)$ (by (\ref{subbasis_gen_by_funcs}) and (\ref{basis_gen_by_funcs})). Now suppose $U \in \mathcal{B}[\mathcal{S}_{\mathcal{D}}(A ;\mathcal{B}_{\rho}(\mathbb{R}))]$ and $y \in U$. Then $U$ has the form
\begin{align}
    U = \bigcap_{i=1}^n \big\{x \in X : \left| a_i - f_i(x) \right| < \epsilon_i \big\} \cap A
\end{align}
for some $f_i \in \mathcal{D}$, $a_i \in \mathbb{R}$, $\epsilon_i > 0$, and $n \in \mathbb{N}$. Since $y \in U$, we have $f_i(y) \in (a_i - \epsilon_i, a_i + \epsilon_i)$ for each $i=1,\ldots,n$. 
Define $\epsilon^*_y \doteq \min_{1 \leq i \leq n} \left\{ \epsilon_i - \left| a_i - f_i(y) \right| \right\}$.
Then
\begin{align}
    V_y &= \bigcap_{i=1}^n \big\{x \in X : \left| f_i(y) - f_i(x) \right| < \epsilon^*_y \big\} \cap A= \big\{x \in A :  \max_{1 \leq i \leq n} \left| f_i(y) - f_i(x) \right| < \epsilon^*_y \big\}
\end{align}
is a basis element of $\mathcal{B}_{\mathcal{D}}(A)$ such that $y \in V_y \subset U$.
\end{proof}

Since we are dealing with general topological spaces, we recall the 
notion of nets.
\begin{definition}[Nets, Subnets, Sequence] \label{def:nets_subnets}
A \textit{net} in a set $X$, denoted $(x_{\lambda})_{\lambda \in \Lambda}$ or simply $(x_{\lambda})$, is a function $P:\Lambda \to X;\lambda\rightarrow x_\lambda$, where $\Lambda$ is some directed set. 

Given a directed set $M$, a \textit{subnet} of $P$ is the composition $P \circ \phi$, where $\phi \colon M \to \Lambda$ satisfies:
\begin{itemize}
    \item $\phi(\mu_1) \leq \phi(\mu_2)$ whenever $\mu_1 \leq \mu_2$, and
    \item for each $\lambda \in \Lambda$, there is some $\mu \in M$ such that $\lambda \leq \phi(\mu)$.
\end{itemize}
A \textit{sequence} is a net whose directed set has the cardinality of the natural numbers.
\end{definition}

\begin{definition}[Net convergence] \label{def:net_convergence}
Let $(x_{\lambda})_{\lambda \in \Lambda}$ be a net in a topological space $X$. Then $(x_{\lambda})_{\lambda \in \Lambda}$ \textit{converges} to $p \in X$ (denoted $x_{\lambda} \to p)$ if and only if for each neighborhood $U$ of $p$, there is some $\lambda_0 \in \Lambda$ such that $\lambda \geq \lambda_0$ implies $x_{\lambda} \in U$. We then say $p$ is a limit of $(x_{\lambda})_{\lambda \in \Lambda}$ and is denoted as $\lim_{\lambda \in \Lambda} x_{\lambda}$ when the limit is unique. 
\end{definition}

Our need for nets is primarily through the following six results.

\begin{proposition} \label{prop:basic_net_properties}
Let $(x_{\lambda})$ be a net in a topological space $X$ and suppose $x \in X$. The following statements are true:
\begin{enumerate}
    \item If $x_{\lambda} = x$ for each $\lambda$, then $x_{\lambda} \to x$.
    \item If $x_{\lambda} \to x$, then every subnet of $(x_{\lambda})$ converges to $x$.
    \item If every subnet of $(x_{\lambda})$ has a subnet converging to $x$, then $(x_{\lambda})$ converges to $x$.
\end{enumerate}
\end{proposition}
\begin{proof}
The first two parts are obvious. 
For (3.), suppose $(x_{\lambda})$ does not converge to $x$. Then, for some neighborhood $U$ of $x$, for each $\lambda$ there is $\lambda_0 \geq \lambda$ such that $x_{\lambda_0} \not \in U$. That is, there is a subnet $(x_{\lambda_{\mu}})$ such that $x_{\lambda_{\mu}} \not \in U$ for all $\mu$. It then follows that a further subnet of $(x_{\lambda_{\mu}})$ cannot converge to $x$, which is a contradiction.
\end{proof}

We can also characterize the closure of a set, that is the smallest closed set
containing that set, in terms of net limits.

\begin{proposition} \label{prop:net_exists_converging_to_any_point_in_closure}
Suppose $X$ is a topological space and $A \subset X$. Then $x \in \overline{A}$ if and only if there is a net $(x_{\lambda})$ in $A$ with $x_{\lambda} \to x$. 
\end{proposition}
\begin{proof}
See \cite{willard} Theorem 11.7.
\end{proof}

For convenience we also list the following eight basics results that we will rely on.

\begin{proposition} \label{prop:continuity_condition_for_nets}
Suppose $X$ and $Y$ are topological spaces and $f \colon X \to Y$. Then $f$ is continuous at $p \in X$ if and only if $f(x_{\lambda}) \to f(p)$ whenever $x_{\lambda} \to p$.
\end{proposition}
\begin{proof}
See \cite{willard} Theorem 11.8.
\end{proof}

\begin{proposition} \label{prop:nets_and_compact_spaces}
A topological space is compact if and only if each net has a convergent subnet.
\end{proposition}
\begin{proof}
See \cite{willard} Theorems 17.4 and 11.5.
\end{proof}

\begin{proposition} \label{prop:homeomorphism_facts}
Suppose $X$ and $Y$ are topological spaces and $f \colon X \to Y$ is a bijection. Then the following are equivalent:
\begin{enumerate}
    \item $f$ is a homeomorphism,
    \item $A \subset X$ is open in $X$ if and only if $f(A)$ is open in $Y$,
    \item $A \subset X$ is closed in $X$ if and only if $f(A)$ is closed in $Y$,
    \item $A \subset X$ implies $f( \mathfrak{cl}[A]) = \mathfrak{cl}[f(A)]$,
    \item For any net $(x_{\lambda})$ and point $p$ in $X$, $x_{\lambda} \to p$ if and only if $f(x_{\lambda}) \to f(p)$.
\end{enumerate}
\end{proposition}
\begin{proof}
See \cite{willard} Theorem 7.9 for (1 - 4). (5) follows from Proposition \ref{prop:continuity_condition_for_nets}.
\end{proof}

\begin{proposition}
Suppose $(x_{\lambda})$ is a net in a Hausdorff space $X$, such that $x_{\lambda} \to p \in X$. Then, $\lim_{\lambda} x_{\lambda} = p$ (i.e. it is the unique limit).
\end{proposition}
\begin{proof}
Obvious. \end{proof}

\begin{definition}[sup metric] \label{def:sup_metric}
If $(Y, d)$ is a metric space and $B(X,Y)$ is the bounded functions from the set $X$ to $Y$, then the \textit{sup metric} is defined on $B(X,Y)$ as 
\begin{align}
    \rho(f,g) = \sup \{d(f(p),g(p)) : p \in X \}.
\end{align}
\end{definition}

\begin{proposition} \label{prop:continuous_bounded_is_closed_complete}
Let $X$ be a topological space and let $(Y,d)$ be a complete metric space. Then $C_B(X,Y)$ is closed and complete in $B(X,Y)$ equipped with the sup metric.
\end{proposition}
\begin{proof}
See \cite{munkres} and explanation on page 270.
\end{proof}

\begin{definition}[Completely Regular, Tychonoff space] \label{def:tychonoff_space}
A topological space $X$ is called $\textit{completely regular}$ if and only if for each $A \in \mathcal{C}(X)$ and point $p \in X \setminus A$ there exists a continuous function $f \colon X \to [0,1]$ such that $f\vert_A=0$ and $f(p)=1$. If $X$ is also Hausdorff, then $X$ is called a \textit{Tychonoff space}.  
\end{definition}

\begin{proposition} \label{prop:subspace_tychonoff}
Subspaces of Tychonoff spaces are Tychonoff.
\end{proposition}
\begin{proof}
See \cite{munkres} Theorem 33.2
\end{proof}

Also, if $(X, \rho)$ is a metric space, then $X$ with the metric topology is a Tychonoff space.


\begin{proposition} \label{prop:product_topology_facts}
Let $X$ be a topological space; $I$ be an index set; $Y_i$ be a topological space for each $i \in I$; $f_i \colon X \to Y_i$ be a mapping for each $i \in I$; and $\mathcal{D} = \{f_i : i \in I\}$. Then the following statements are true:
\begin{enumerate}
   \item $\bigotimes \mathcal{D} \colon X \to \prod_{i \in I} Y_i$ is an embedding if and only if it is injective and $\mathcal{O}(X) = \mathcal{O}_{\mathcal{D} }(X)$. 
    \item A net $(x_{\lambda})$ in $\prod_{i \in I} Y_i$ converges to $p$ if and only if for each $i \in I$, $\pi_i(x_{\lambda}) \to \pi_i(p)$ in $Y_i$. 
    \item If $Y_i$ is Hausdorff for each $i \in I$, then $\prod_{i \in I} Y_i$ is Hausdorff.
    \item If $Y_i$ is Tychonoff for each $i \in I$, then $\prod_{i \in I} Y_i$ is Tychonoff.
\end{enumerate}
\end{proposition}
\begin{proof}
See \cite{munkres} Theorems 18.1 and 19.6 for (1-3), 19.4 for (3), and 33.2 for (4). See \cite{willard} Theorem 8.12 for (1) and 11.9 for (2).
\end{proof}


\subsection{Sequential Spaces} \label{sec:sequential_spaces}

Sequential spaces are topological spaces whose properties can be established
using sequence subnets, rather than having to deal with general nets. 
A sequential space can be generated from any topology. 
The generated sequential space shares the same convergent sequences as the original space. 

\begin{definition}[Eventually in, Sequentially Open, Sequential Space] \label{def:sequential_space}
Suppose $X$ is a topological space, $A \subset X$, and $(x_n)$ is a sequence in $X$. We say $(x_n)$ is \textit{eventually in} $A$ if there is an $N \in \mathbb{N}$ such that $n \geq N$ implies $x_n \in A$. The set $A$ is said to be \textit{sequentially open} if every sequence in $X$ that converges to a point in $A$ is eventually in $A$. $X$ is a \textit{sequential space} if every sequentially open set is open.
\end{definition}

\begin{definition}[Sequentially Continuous]
Given topological spaces $X$ and $Y$, a function $f \colon X \to Y$ is \textit{sequentially continuous} if for any sequence $(x_n)$ and point $p \in X$ such that $x_n \to p$ we have $f(x_n) \to f(p)$.
\end{definition}

The next two results are obvious.

\begin{proposition} \label{prop:open_is_sequential_open}
Open sets are sequentially open.
\end{proposition}

\begin{proposition} \label{prop:sequential_space_sequential_continuous}
Let $X$ be a sequential space and Y be a topological space. Then $f \colon X \to Y$ is continuous if and only if $f$ is sequentially continuous.
\end{proposition}

The collection of sequentially open sets defines a finer topology than the original space. 

\begin{proposition} \label{prop:gen_seq_top}
Suppose $(X, \mathcal{T})$ is a topological space. The collection of sequentially open sets, denoted $\mathcal{T}_s$, is a topology on $X$. Further $\mathcal{T} \subset \mathcal{T}_s$.
\end{proposition}
\begin{proof}
First, we show $\mathcal{T}_s$ is a topology on $X$. Clearly, $\{\emptyset, X\} \subset \mathcal{T}_s$. Suppose $\mathcal{C} \subset \mathcal{T}_s$ and let $A_{\mathcal{C}} = \bigcup_{A \in \mathcal{C}} A$. If $p \in A_{\mathcal{C}}$, then there is some $A_0 \in \mathcal{C}$ such that $p \in A_0$. Therefore if $(x_n)$ converges to $p$, then it is eventually in $A_0$ and $A_{\mathcal{C}}$. Next, suppose $\mathcal{C}_0 \in \mathcal{R}_0(\mathcal{T}_s)$ and let $A_{\mathcal{C}_0} = \bigcap_{A \in \mathcal{C}_0} A$. If $p \in A_{\mathcal{C}_0}$, then $p \in A$ for each $A \in \mathcal{C}_0$. Therefore if $(x_n)$ converges to $p$, then we can define $N_A$ such that $x_n \in A$ for each $n \geq N_A$ and choose $N = \max\{N_A : A \in \mathcal{C}_0\}$ implying $x_n \in A_{\mathcal{C}_0}$ when $n \geq N$.

Open sets are sequentially open by Proposition \ref{prop:open_is_sequential_open}, so $\mathcal{T} \subset \mathcal{T}_s$.
\end{proof}

\begin{definition}[Generating a sequential space] \label{def:gen_seq_top}
Given a topological space $(X, \mathcal{T})$, $\mathfrak{Q}(X, \mathcal{T})$ denotes the sequential topology on $X$ generated from $\mathcal{T}$. 
\end{definition}
A topological space $(X, \mathcal{T})$ is a sequential space if and only if $\mathcal{T} = \mathfrak{Q}(X, \mathcal{T})$.
The next result shows that a topological space $X$ and its generated sequential space share the same convergent sequences. 

\begin{proposition} \label{prop:same_convergent_sequences}
Suppose $(X, \mathcal{T})$ is a topological space and let $\mathcal{T}_s = \mathfrak{Q}(X, \mathcal{T})$. Then, a sequence $(x_n)$ converges to $p \in X$ with $ \mathcal{T}$ if and only if it converges to $p$ with $\mathcal{T}_s$.
\end{proposition}
\begin{proof}
$\mathcal{T} \subset \mathcal{T}_s$ by Proposition \ref{prop:gen_seq_top} so convergence with $\mathcal{T}_s$ implies convergence with $\mathcal{T}$. 
Conversely, suppose $(x_n)$ converges to $p$ with $\mathcal{T}$ and $A_p \in \mathcal{T}_s$ satisfies $p\in A_p$. $A_p $ is sequentially open with respect to $\mathcal{T}$ so $(x_n)$ is eventually in $A_p$. Therefore, $x_n \to p$ with $\mathcal{T}_s$.
\end{proof}

Every sequence is a net, so sequential limit points are always limit points.

\begin{definition}[Sequential Limit Point]
Let $X$ be a topological space and $A \subset X$. Then, $p$ is a \textit{sequential limit point} of $A$ if there exists a sequence in $A$ converging to $p$.
\end{definition}

\begin{proposition} \label{prop:only_squences_when_first_countable}
Let $X$ be a first-countable topological space. Then,
\begin{enumerate}
    \item A point $p \in X$ is a limit point of $A \subset X$ if and only if $p$ is a sequential limit point of $A$.
    \item $X$ is a sequential space.
\end{enumerate}
\end{proposition}
\begin{proof}
See \cite{munkres} Theorem 30.1 for (1.). 
We show (1.) implies sequentially open sets are open. 
Let $B \subset X$ be sequentially open and $p$ be a limit point of $X \setminus B$ (if any), so by (1.) there must be a sequence $(x_n)$ in $X \setminus B$ converging to $p \in X$. If $p \in B$, then $(x_n)$ must be eventually in $B$; however, $x_n \in X \setminus B$ for all $n$, so a contradiction has been reached. Therefore, $X \setminus B$ contains all of its limit points, which is to say that it is closed.  
\end{proof}
Often we work with a metric space in universal approximation so Propositions 
\ref{prop:metrizable_is_fist_countable} and \ref{prop:only_squences_when_first_countable} tell us we can check topological properties using sequences instead of nets. 

\begin{proposition} \label{prop:metrizable_is_fist_countable}
Metrizable spaces are first-countable.
\end{proposition}
\begin{proof}
See \cite{munkres} pages 130/131.
\end{proof}

Our universal approximation results rely on the construction of homeomorphisms, so it is good to know that homeomorphisms preserve sequential spaces. 

\begin{proposition}
Suppose $(X, \mathcal{T}_X)$ is a sequential space and $(Y, \mathcal{T}_Y)$ is a topological space. If $h \colon X \to Y$ is a homeomorphism, then $Y$ is a sequential space.
\end{proposition}
\begin{proof}
Suppose $A \subset Y$ is sequentially open. We must show $A \in \mathcal{T}_Y$. Let $(x_n)$ be a sequence converging to $p \in f^{-1}(A)$. 
By Proposition \ref{prop:homeomorphism_facts} (5), it follows then that $f(x_n)$ is a sequence converging to $f(p) \in A$, so $f(x_n)$ is eventually in $A$ since $A$ is sequentially open. 
Therefore, there is some $N \in \mathbb{N}$ such that $x_n \in f^{-1}(A)$ when $n \geq N$. 
That is, $x_n$ is eventually in $f^{-1}(A)$, so $f^{-1}(A)$ is sequentially open and so it is open (as $X$ is a sequential space). It then follows from Proposition \ref{prop:homeomorphism_facts} (2) that $A$ is open.
\end{proof}

The next result is important for homeomorphism construction as it reduces continuity in the reverse direction to sequential continuity.

\begin{proposition} \label{prop:sequential_continuous_inverse}
Suppose $X$ is a topological space; $\mathcal{M} \subset \mathbb{R}^X$ is countable; and $\bigotimes \mathcal{M} \colon X \to \bigotimes \mathcal{M}(X)$ is a bijection. Then $[\bigotimes \mathcal{M}]^{-1}$ is continuous if and only if, for any sequence $(x_n)$ in $X$, we have $g(x_n) \to g(p)$ for all $g \in \mathcal{M}$ implies $x_n \to p$ in $X$. 
\end{proposition}
\begin{proof}
Since $\mathcal{M}$ is countable, $\mathbb{R}^{\mathcal{M}}$ is metrizable and so is $\bigotimes \mathcal{M}(X)$ as a subspace, so it is first-countable by Proposition \ref{prop:metrizable_is_fist_countable}. 
It then follows from Propositions \ref{prop:only_squences_when_first_countable} and \ref{prop:sequential_space_sequential_continuous} that we only need $[\bigotimes \mathcal{M}]^{-1}$ to be sequentially continuous for it to be continuous. 
Suppose $(y_n)$ is a sequence in $\bigotimes \mathcal{M}(X)$ converging to $q$ and define $x_n \doteq [\bigotimes \mathcal{M}]^{-1}(y_n)$, $p \doteq [\bigotimes \mathcal{M}]^{-1}(q)$ in $X$. 
However, $(y_n)$ converges to $q$ in $\bigotimes \mathcal{M}(X)$ if and only if $(\pi_g(y_n))=(g(x_n))$ converges to $\pi_g(q)=g(p)$ for each $g \in \mathcal{M}$. 
Therefore, by hypothesis $y_n \to q$ in $\bigotimes \mathcal{M}(X)$ implies $x_n \to p$ in $X$ and $[\bigotimes \mathcal{M}]^{-1}$ is sequentially continuous and hence is continuous.
\end{proof}

\subsection{Point Separation}

If $X$ is a topological space, then we wish to identify collections of $\mathbb{R}$-valued functions $\mathcal{M}$ such that $\bigotimes \mathcal{M} \colon X \to \mathbb{R}^{\mathcal{M}}$ is an embedding into a compact subset of $\mathbb{R}^{\mathcal{M}}$.
Our goal is achieved when $\mathcal{M}$ both separate points (s.p.) and strongly separate points (s.s.p.) on $X$. 
When $\mathcal{M}$ is countable the s.s.p. condition can be confirmed by checking sequential continuity of $[\bigotimes \mathcal{M}]^{-1}$.

\begin{proposition} 
\label{lem:sep_or_spp_may_imply_hausdorff}
Let $X$ be a topological space and $\mathcal{M} \subset \mathbb{R}^X$ s.p. on $X$. Then,
\begin{enumerate}
    \item $\bigotimes \mathcal{M} \colon X \to \bigotimes \mathcal{M}(X)$ is a bijection,
    \item $\mathcal{M} \subset C(X)$ implies $X$ is a Hausdorff space,
    \item $\mathcal{M} \subset \mathcal{N} \subset \mathbb{R}^X$ implies $\mathcal{N}$ s.p. on $X$, and
    \item $A \subset X$ implies $\mathcal{M}\vert_A$ s.p. on $A$.
\end{enumerate}

\end{proposition}
\begin{proof}
(1.) Suppose $p \neq q \in X$. $\mathcal{M}$ s.p. on $X$ implies there is some $g \in \mathcal{M}$ such that $g(p) \neq g(q)$ so $\bigotimes \mathcal{M}(p) \neq  \bigotimes \mathcal{M}(q)$. Hence, for each $r \in \bigotimes \mathcal{M}(X)$ there is a unique $p \in X$ such that $\bigotimes \mathcal{M}(p) = r$ and $\bigotimes \mathcal{M}$ is a bijection.

(2.)
For $p\ne q \in X$, there is a $f\in\mathcal M\subset C(X)$ such that  $r = \frac{|f(p) - f(q)|}{3} > 0$. 
$(f(p) - r, f(p) + r)$ and $(f(q) - r, f(q) + r)$ are disjoint open balls containing $f(p)$ and $f(q)$, respectively. Since $f$ is continuous, ${f^{-1}[(f(p) - r, f(p) + r)] \ni p}$ and ${f^{-1}[(f(q) - r, f(q) + r)] \ni q}$ are open in $X$ with empty intersection. It then follows that $X$ is Hausdorff.

(3.) and (4.) are obvious.
\end{proof}

The following result is obvious.

\begin{lemma} \label{lem:ssp_properties}
Let $X$ be a topological space; $\mathcal{M} \subset \mathbb{R}^X$ be a collection of real valued functions on $X$; and $A$ be a subspace of $X$. The following properties hold:
\begin{enumerate}
    \item If $\mathcal{M}$ s.s.p. and $\mathcal{M} \subset \mathcal{N} \subset \mathbb{R}^X$, then $\mathcal{N}$ also s.s.p. \label{lem:ssp_property1}
    \item If $\mathcal{M}$ s.s.p. on $X$, then $\mathcal{M}\vert_A$ s.s.p. on $A$.  \label{lem:ssp_property2}
    \item Let $\mathcal{T}_1 \subset \mathcal{T}_2$ be topologies on $X$. If $\mathcal{M}$ s.s.p. on $(X, \mathcal{T}_2)$, then $\mathcal{M}$ s.s.p. on $(X, \mathcal{T}_1)$. \label{lem:ssp_property3}
\end{enumerate}
\end{lemma}

There is an alternative means of defining the s.s.p. property, which can be useful when the topology of a space is defined by a collection of real valued functions like in Definition \ref{def:toplogy_define_by_functions}.

\begin{proposition} \label{prop:ssp_equiv_to_finer_topology}
Suppose $(X, \mathcal{T})$ is a topological space and $\mathcal{M} \subset \mathbb{R}^X$ is a collection of real valued functions on $X$. Then $\mathcal{M}$ s.s.p. on $(X, \mathcal{T})$ if and only if $\mathcal{T} \subset \mathcal{O}_{\mathcal{M}}(X)$.
\end{proposition}
\begin{proof}
By Proposition \ref{prop:basis_for_topology_generated_by_functions}, $\mathcal{B}_\mathcal{M}(X)$ is a basis for $\mathcal{O}_{\mathcal{M}}(X)$ with sets of the form
\begin{align}
     B_{q, \epsilon}(\mathcal{M}_0) \doteq \big\{p \in X : \max_{f \in \mathcal{M}_0} \left| f(q) - f(p) \right| < \epsilon \big\}  && \mathcal{M}_0 \in \mathcal{R}_0 (\mathcal{M}).
\end{align}
If $\mathcal{M}$ s.s.p., then for each $O \in \mathcal{T}$ there exist $q$, $\mathcal{M}_0 \in \mathcal{R}_0 (\mathcal{M})$ and $\epsilon > 0$ such that $B_{q, \epsilon}(\mathcal{M}_0) \subset O$. It follows that $\mathcal{T} \subset \mathcal{O}_{\mathcal{M}}(X)$.

Conversely, assume $\mathcal{T} \subset \mathcal{O}_{\mathcal{M}}(X)$. For each neighborhood $N_q$ of $q$, there is an $O_q \in \mathcal{T}$ such that $q \in O_q \subset N_q$. By assumption, $O_q \in \mathcal{T}$ implies $O_q \in \mathcal{O}_{\mathcal{M}}(X)$, so there is $\mathcal{M}_0 \in \mathcal{R}_0 (\mathcal{M})$ and $\epsilon > 0$ such that $B_{q, \epsilon}(\mathcal{M}_0) \subset O_q$. It then follows that $ \inf_{p \not \in O_q} \{\max_{f \in \mathcal{M}_0} \left| f(q) - f(p) \right| \} \geq \epsilon > 0$, so $\mathcal{M}$ s.s.p.
\end{proof}

Now we show s.p. is implied by s.s.p. and the Hausdorff property.

\begin{proposition}   \label{lem:spp_implies_sp_on_hausdorff}
Let $X$ be a topological space, $A \subset X$ be non-empty, and $\mathcal{D} \subset \mathbb{R}^X$. Then, the following statements are true:

\begin{enumerate}
\item[(a)] If $\{ \{x\}: x \in A\} \subset \mathcal{C}(X)$, especially if $A$ is a Hausdorff subspace of $X$, then $\mathcal{D}$ strongly separating points on $A$ implies $\mathcal{D}$ separating points on $A$. 
\item[(b)] $\mathcal{D}$ separates points on $A$ if and only if $(A, \mathcal{O}_{\mathcal{D}}(A))$ is a Hausdorff space.
\end{enumerate}
\end{proposition}
\begin{proof}
We provide the proof from \cite{dong_kouritzin} Proposition 9.2.1.

(a) The Hausdorff property of $(A, \mathcal{O}_X(A))$ implies $\{ \{x\}: x \in A\} \subset \mathcal{C}(A, \mathcal{O}_X(A))$. 
Now, if $\mathcal{D}$ fails to s.p. on $A$, then there are $x\ne y \in A$ such that $\bigotimes \mathcal{D}(x) = \bigotimes \mathcal{D}(y)$. Since $\{y\}$ is closed and $\mathcal{D}$ strongly separates points on $A$, there exist $\mathcal{D}_x \in \mathcal{R}_0(\mathcal{D})$ and $\epsilon \in (0, \infty)$ such that $y \in \{ z \in A : \max_{f \in \mathcal{D}_x} \left| f(x) - f(z) \right|  < \epsilon \} \subset A \setminus \{y\}$, which is a contradiction.

(b - Sufficiency) follows by (a) (with $A = (X, \mathcal{O}_{\mathcal{D}}(X))$).

(b - Necessity) Let $x_1\ne x_2 \in A$. Since $\mathcal{D}$ s.p. on $A$, there is an $f \in \mathcal{D}$ such that $\epsilon_0 \doteq \left| f(x_1) - f(x_2) \right| >0$. Letting $O_i \doteq \{ z \in A : \left| f(x_i) - f(z) \right| < \frac{\epsilon_0}{3} \} \in \mathcal{O}_{\mathcal{D}}(A)$ for $i = 1,2$, we have $x_1 \in O_1$, $x_2 \in O_2$ and $O_1 \cap O_2 = \emptyset$.
\end{proof}

The next result shows continuous functions that s.s.p. and s.p. only occur in Tychonoff spaces, which is why our universal approximation results are limited to Tychonoff spaces. 

\begin{proposition}
\label{lem:tychonoff_equiv_sep_and_ssp}
Let $X$ be a topological space. Then, the following are equivalent:
\begin{enumerate}
    \item $X$ is a Tychonoff space.
    \item $C(X)$ separates and strongly separates points on $X$.
    \item $C_B(X)$ separates and strongly separates points on $X$.
\end{enumerate}
\end{proposition}
\begin{remark}
If $\mathcal{M}\subset C(X)$ s.p and s.s.p., then $C(X)$ s.p. and s.s.p. so $X$ is Tychonoff.
\end{remark}
\begin{proof}
We provide the proof from \cite{dong_kouritzin} Proposition 9.3.1.

($1 \to 2$) Suppose $O_p \in  \mathcal{O}(X)$ is an open neighborhood of $p \in X$ and let $A = X \setminus O_p \in \mathcal{C}(X)$. 
Since $E$ is a Tychonoff space, there is an $f_{A,p} \in C(X;[0,1])$ such that $f_{A,p} \vert_A=0$ and $f_{A,p}(p)=1$ by Definition \ref{def:tychonoff_space}. It follows that $\{q \in X : | f_{A,p}(p) - f_{A,p}(q) | < \epsilon  \} \subset O_p$ for each $\epsilon \in (0,1)$ so $\mathcal{O}(X) \subset  \mathcal{O}_{C(X)}(X)$, which implies $C(X)$ s.s.p. on $X$ by Proposition \ref{prop:ssp_equiv_to_finer_topology}. Proposition \ref{lem:spp_implies_sp_on_hausdorff} (a) implies $C(X)$ s.p. on $X$.

($2 \to 3$) 
Let $\rho$ be the standard metric on $\mathbb{R}$.
For any collection of $\mathbb{R}$-valued functions $\mathcal D$ on $X$, ${\mathcal{S}_{\mathcal{D}}(X ; \mathcal{B}_{\rho}(\mathbb{R})) = \left\{ \{q \in X : a - r < f(q) < a + r \}  : a \in \mathbb{R}, r > 0 , f \in \mathcal{D} \right\}}$ is a subbasis for $\mathcal{O}_{\mathcal{D}}(X)$ by Proposition \ref{prop:subbasis_for_O_D}. 
For each $f \in C(X)$, define the function \\ $g_{r,a,f} \doteq  (f \vee (a - r)) \wedge ( a + r)$, which is bounded, continuous, and satisfies $\{q \in X : a - r < g_{r,a,f}(q) < a + r \} = \{q \in X : a - r < f(q) < a + r \}$. So $\mathcal{O}_{C_B(X)}(X) \supset \mathcal{O}_{C(X)}(X) \supset \mathcal{O}(X)$; and hence, $C_B(X)$ s.s.p. on $X$ by Proposition \ref{prop:ssp_equiv_to_finer_topology}. By Proposition \ref{lem:sep_or_spp_may_imply_hausdorff}, $X$ is Hausdorff, so Proposition \ref{lem:spp_implies_sp_on_hausdorff} (a) implies $C_B(X)$ s.p. on $X$.

($3 \to 1$) Pick $p \in X$ and $A \in \mathcal{C}(X)$ such that $p \not \in A$. Since $C_B(X)$ s.s.p. on $X$, there exist $\mathcal{M}_0 \in \mathcal{R}_0(C_B(X))$ and $\epsilon > 0$ such that 
\begin{align}
    {p \in \left\{q \in X : \max_{f \in \mathcal{M}_0} | f(p) - f(q) | < \epsilon  \right\} \subset X \setminus A},
\end{align}
from which it follows that
\begin{align}
    h(q) \doteq 1 - \min \left\{1, \frac{\max_{f \in \mathcal{M}_0} \left|  f(p) - f(q)  \right|}{\epsilon} \right\}
\end{align}
is a continuous function from $X$ to $[0,1]$ such that $h \vert_A = 0$ and $h(p) =1$. Hence, by Definition \ref{def:tychonoff_space}, $X$ is a Tychonoff space.
\end{proof}

s.s.p. is intimately related to another important property.

\begin{definition}[Determines Point Convergence, Determines Sequential Point Convergence] \label{def:determins_point_convergence}
Let $X$ be a topological space and $\mathcal{M} \subset \mathbb{R}^X$. We say $\mathcal{M}$ \textit{determines point convergence on} $X$ if and only if, for any net $(x_\lambda)$ and point $p$ in $X$, we have $g(x_{\lambda}) \to g(p)$ for each $g \in \mathcal{M}$ implies $x_{\lambda} \to p$ in $X$. Similarly, we say $\mathcal{M}$ \textit{determines sequential point convergence on} $X$ if and only if, for any sequence $(x_n)$ and point $p$ in $X$, we have $g(x_n) \to g(p)$ for each $g \in \mathcal{M}$ implies $x_n \to p$ in $X$.
\end{definition}

\begin{proposition}
Let $X$ be a topological space and $\mathcal{M} \subset \mathbb{R}^X$. Then, $\mathcal{M}$ determines point convergence on $X$ implies $\mathcal{M}$ determines sequential point convergence on $X$.
\end{proposition}
\begin{proof}
By Definition \ref{def:nets_subnets}, sequences are nets. The rest is obvious.
\end{proof}

The next result relates homeomorphisms to s.s.p. and determining point convergence. 

\begin{proposition} \label{prop:ssp_homeomorphism_determine_point_convergence}
Suppose $X$ be a Hausdorff space or $C(X)$ s.p. on $X$ and let $\mathcal{M} \subset \mathbb{R}^X$. Statements (1 - 3) are equivalent and imply (4). If $\mathcal{M}$ is countable then (1 - 4) are equivalent.
\begin{enumerate}
    \item $\bigotimes \mathcal{M} \colon X \to \bigotimes \mathcal{M}(X)$ has a continuous inverse. If $\mathcal{M} \subset C(X)$, then $\bigotimes \mathcal{M} \colon X \to \bigotimes \mathcal{M}(X)$ is a homeomorphism.
    \item $\mathcal{M}$ s.s.p. on $X$.
    \item $\mathcal{M}$ determines point convergence on $X$.
    \item $\mathcal{M}$ determines sequential point convergence on $X$.
\end{enumerate}
\end{proposition}
\begin{proof}
$(1 \leftrightarrow 2)$ See \cite{blount} Lemma 1.

$(2 \leftrightarrow 3)$ See \cite{blount} Lemma 4.

$(3 \to 4)$ Follows directly from Definition \ref{def:determins_point_convergence} as all sequences are nets.

$(4 \to 1$, $\mathcal{M}$ countable) Follows from Proposition \ref{prop:sequential_continuous_inverse}.
\end{proof}

There is a simple a condition for a \emph{countable} subcollection of functions that s.s.p. 

\begin{proposition} \label{prop:countable_when_countable_basis}
If $(X, \mathcal{T})$ has a countable basis and $\mathcal{M} \subset C(X)$ s.s.p., then there is a countable
collection $\{g_i\}_{i=0}^{\infty} \subset \mathcal{M}$  that s.s.p. Moreover, $\{g_i\}_{i=0}^{\infty}$ can be taken closed under either
multiplication or addition if $\mathcal{M}$ is.
\end{proposition}
\begin{proof}
Proven by \cite{blount} Lemma 2.
\end{proof}

Recall that Proposition \ref{lem:tychonoff_equiv_sep_and_ssp} tells us the bounded continuous functions s.s.p. on Tychonoff spaces. Hence, a Tychonoff space with a countable base ensures there is a countable collection of bounded continuous functions that are closed under multiplication and s.s.p.

\begin{proposition}
\label{lem:compact_hausdorff_spp_equiv_sp}
Let $X$ be a compact space and $\mathcal{M} \subset C(X)$. Then, $X$ is a Hausdorff space and $\mathcal{M}$ strongly separates points on $X$ if and only if $\mathcal{M}$ separates points on $X$. 
\end{proposition}

\begin{proof}
Due to Proposition \ref{lem:sep_or_spp_may_imply_hausdorff} (2) and Proposition \ref{lem:spp_implies_sp_on_hausdorff} (a), we need only show $\mathcal{M}$ s.p. on the compact $X$ implies $\mathcal{M}$ s.s.p. on $X$. Further, by Proposition \ref{prop:ssp_homeomorphism_determine_point_convergence}, we need only show $\mathcal{M}$ s.p. implies $\mathcal{M}$ determines point convergence on $X$.

Let $(x_{\lambda})$ be a net such that $\lim_{\lambda} f(x_{\lambda}) = f(x)$ for all $f \in \mathcal{M}$. 
By compactness and Proposition \ref{prop:nets_and_compact_spaces} there exists a subnet $(x_{\lambda_{\mu}})$ and $p \in X$ such that $x_{\lambda_{\mu}} \to p$.
Hence, for each $f \in \mathcal{M}$, $\lim_{\mu} f(x_{\lambda_{\mu}}) = f(x)$ and, since $x_{\lambda_{\mu}} \to p$ and $f$ is continuous, we also have ${\lim_{\mu} f(x_{\lambda_{\mu}}) = f(x) = f(p)}$. As $\mathcal{M}$ s.p. on $X$, we have $x = p$, so $x_{\lambda_{\mu}} \to x$. 
As every subnet has a subnet converging to $x$, we have $x_{\lambda} \to x$, so $\mathcal{M}$ determines point convergence and s.s.p. on $X$.
\end{proof}

So when $X$ is compact, it is enough to show $\mathcal{M} \subset C(X)$ s.p. on $X$ to get the s.s.p.

The following result can be useful in identifying when a collection s.s.p.

\begin{proposition}
\label{lem:dense_sets_still_sp_or_spp}
Let $(X, \mathcal{T})$ be a topological space and the members of $\mathcal{M}_0 \subset C(X)$ and $\mathcal{M} \subset \mathbb{R}^X$ be bounded. 
Suppose $\mathcal{M} \subset \overline{\mathcal{M}}_0$ (where the bar denotes closure under the sup metric). If $\mathcal{M}$ separates points or strongly separates points, then $\mathcal{M}_0$ does also.
\end{proposition}

\begin{proof}
$\overline{\mathcal{M}}_0  \subset C_B(X, \mathcal{O}_{\mathcal{M}_0}(X); \mathbb{R})$ by Proposition \ref{prop:continuous_bounded_is_closed_complete} so $\mathcal{M}  \subset C(X, \mathcal{O}_{\mathcal{M}_0}(X); \mathbb{R})$ and $\mathcal{O}_{\mathcal{M}}(X) \subset \mathcal{O}_{\mathcal{M}_0}(X)$. 
Now, $\mathcal{M}$ s.p. on $(X, \mathcal{T})$ implies $(X, \mathcal{O}_{\mathcal{M}}(X))$ is Hausdorff by Proposition \ref{lem:spp_implies_sp_on_hausdorff} (b). Then, $\mathcal{O}_{\mathcal{M}}(X) \subset \mathcal{O}_{\mathcal{M}_0}(X)$ implies $(X, \mathcal{O}_{\mathcal{M}_0}(X))$ is also a Hausdorff space so $\mathcal{M}_0$ s.p. on $(X, \mathcal{T})$ too by Proposition \ref{lem:spp_implies_sp_on_hausdorff}.
Finally, $\mathcal{M}$ s.s.p. on $(X, \mathcal{T})$ implies $\mathcal{T} \subset \mathcal{O}_{\mathcal{M}}(X) \subset \mathcal{O}_{\mathcal{M}_0}(X)$ by Proposition \ref{prop:ssp_equiv_to_finer_topology}, so $\mathcal{M}_0$ s.s.p. on $(X, \mathcal{T})$ too.
\end{proof}

\subsection{Uniform Spaces}

We now discuss uniform spaces and uniformly continuous functions
on (not necessarily metric) uniform spaces.
Both of the following metrics on $\mathbb{R}$:
\begin{align}
    d_1(x,y) \doteq \left| x - y \right| && d_2(x,y) \doteq \left| \arctan x - \arctan y \right|
\end{align}
generate the standard topology. 
The continuous function $f \colon (\mathbb{R}, d_2) \to (\mathbb{R}, d_1)$ defined as $f(x) \doteq x$ for $x\in \mathbb{R}$ is not uniformly continuous.
But, $f \colon (\mathbb{R}, d_i) \to (\mathbb{R}, d_i)$ for each $i=1,2$ are
so uniform continuity is not just about the topologies of a function's domain and range. 
Uniform spaces and uniformities are used to extend uniform continuity to spaces without a metric.  (See sections 35-37 of \cite{willard}.)

Uniform continuity requires a notion of closeness between points. 
If there is a metric $d$, then the set 
$\{(x,y) : d(x,y) < \epsilon; \, x,y \in X \}$ consists of point pairs within $\epsilon$ of each other. 
Otherwise, we extend the definition of closeness beyond metric spaces 
by considering subsets of the Cartesian product $X \times X$ with the 
following notation:
If $X$ is a set, then the \textit{diagonal} of $X$ is $\Delta(X) \doteq \{(x,x) : x \in X\}$.
If $A$ is a subset of $X \times X$, then $A$ is called a \textit{relation}. If $(x,y) \in A$, then $x$ is \textit{related} to $y$. 
If $x$ is related to $y$ it is not necessarily the case that $y$ is related to $x$. 
If $A$ is a relation, then the \textit{inverse relation} of $A$ is defined as $A^{-1} \doteq \{(y,x) \colon (x,y) \in A\}$. If $A = A^{-1}$, then $A$ is \textit{symmetric}.
 If $A$ and $B$ are subsets of $X \times X$, then the \textit{composition} of $A$ and $B$ is defined as $A \circ B \doteq \{(x,y) : \text{for some } z \in X, (x,z) \in A \text{ and } (z,y) \in B \}$.

The definitions of a uniformity and a uniform space are as follows.

\begin{definition}[Uniformity]
A \textit{uniformity}\footnote{There are equivalent definitions of a uniformity as explained by \cite{willard} chapters 35 and 36. The definition used in this document is for diagonal uniformities.} on a set $X$ is a collection $\mathfrak{D}$ of subsets of $X \times X$ which satisfy:
        \begin{enumerate}
            \item $A \in \mathfrak{D} \Longrightarrow \Delta(X) \subset A$, \label{uniformity_a}
            \item $A_1, A_2 \in \mathfrak{D} \Longrightarrow A_1 \cap A_2 \in \mathfrak{D}$, \label{uniformity_b}
            \item $A \in \mathfrak{D} \Longrightarrow B \circ B \subset A$ for some $B \in \mathfrak{D}$, \label{uniformity_c}
            \item $A \in \mathfrak{D} \Longrightarrow B^{-1} \subset A$ for some $B \in \mathfrak{D}$, and \label{uniformity_d}
            \item $A \in \mathfrak{D}$ and $A \subset B \subset X \times X \Longrightarrow B \in \mathfrak{D}$. \label{uniformity_e}
        \end{enumerate}
\end{definition}

\begin{definition}[Uniform Space, Surroundings]
A set $X$ together with uniformity $\mathfrak{D}$ form a \textit{uniform space}. The members of $\mathfrak{D}$ are called \textit{surroundings}. Given a uniform space $X$, we use the notation $\mathcal{U}(X)$ to denote the uniformity on $X$. 
\end{definition}

Looking at the definition of a uniformity, we can see the remnants of a metric. Condition \ref{uniformity_a} is an extension of $d(x,y) = 0$ if and only if $x = y$, condition \ref{uniformity_c} comes from the triangle inequality (see accompanying proof for Definition \ref{def:metric_uniformity}), and condition \ref{uniformity_d} is analogous to the symmetry of a metric ($d(x,y) = d(y,x)$). 

\begin{proposition} \label{prop:uniform_subspace}
Suppose $X$ is a uniform space and $A \subset X$. The collection defined as $\mathcal{U}_X(A) \doteq \{ D \cap (A \times A) : D \in \mathcal{U}(X) \}$ is a uniformity on $A$.
\end{proposition}
\begin{proof}
For each $E \in \mathcal{U}_X(A)$, there is a $D \in \mathcal{U}(X)$ so that $E = D \cap (A \times A)$.\vspace*{0.1cm}\\
\ref{uniformity_a}.\ \ $\Delta(X) \subset D$ implies $\Delta(A)=\Delta(X) \cap (A \times A) \subset D \cap (A \times A)=E$ for each $E \in \mathcal{U}_X(A)$.\vspace*{0.1cm}\\
\ref{uniformity_b}.\ \ Suppose $D_i \in \mathcal{U}(X)$, $E_i \in \mathcal{U}_X(A)$, and $E_i = D_i \cap (A \times A)$ for $i=1,2$. Then $E_1 \cap E_2 = D_1 \cap D_2 \cap (A \times A)$, and $D_1 \cap D_2 \in \mathcal{U}(X)$, so $E_1 \cap E_2 \in \mathcal{U}_X(A)$.\vspace*{0.1cm}\\
\ref{uniformity_c}.\ \  $D \in \mathcal{U}(X)$ implies a $B \in \mathcal{U}(X)$ such that $B \circ B \subset D$. 
Suppose $(a,b) \in \left[B \cap (A \times A)\right] \circ \left[B \cap (A \times A)\right]$. Then there exists a $z^* \in X$ such that $(a, z^*) \in B \cap (A \times A) $ and $(z^*, b) \in B \cap (A \times A)$. However, $B \cap (A \times A) \subset B$. Therefore, $(a,b) \in B \circ B$ and $(a,b) \in (A \times A) \circ (A \times A) = A \times A$ so $\left[B \cap (A \times A)\right] \circ \left[B \cap (A \times A)\right] \subset B \circ B \cap (A \times A) \subset D \cap (A \times A)$.\vspace*{0.1cm}\\
\ref{uniformity_d}.\ \  $D \in \mathcal{U}(X)$ implies a $B \in \mathcal{U}(X)$ such that $B^{-1} \subset D$. 
Both $D \cap (A \times A)$ and $B \cap (A \times A)$ are elements of $\mathcal{U}_X(A)$, so $[B \cap (A \times A)]^{-1} = B^{-1} \cap (A \times A) \subset D \cap (A \times A)$.\vspace*{0.1cm}\\
\ref{uniformity_e}.\ \  Suppose $E_1 \subset E_2 \subset A \times A$ and $E_1 \in \mathcal{U}_X(A)$. Then, there is a $D \in \mathcal{U}(X)$ such that $E_1 = D \cap (A \times A)$. Clearly, $D \subset D \cup E_2$, so $D \cup E_2 \in \mathcal{U}(X)$, and it follows that $(D \cup E_2) \cap (A \times A) \in \mathcal{U}_X(A)$. However, $(D \cup E_2) \cap (A \times A) = [E_2 \cap (A \times A)] \cup [D \cap (A \times A)] = E_2$.
\end{proof}

\begin{definition}[Relative Uniformity, Uniform Subspace]
Suppose $X$ is a uniform space and $A \subset X$. Then, $\mathcal{U}_X(A)$ (defined in Proposition \ref{prop:uniform_subspace}) is called the \textit{relative uniformity} induced on $A$ by $X$. With this uniformity, $A$ is called a \textit{uniform subspace} of $X$.
\end{definition}

Like how a topological space may be generated from a topological basis, we can generate a uniformity from a uniform basis. 

\begin{definition}[Uniform Base]
$\mathfrak{E}$ is a \textit{uniform base} for $\mathfrak{D}$ if and only if $\mathfrak{E} \subset \mathfrak{D}$ and each $D \in \mathfrak{D}$ contains some $E \in \mathfrak{E}$. A uniformity is generated from a base through repeated use of condition \ref{uniformity_e}; that is $\mathfrak{D} = \{D \subset X \times X \mid E \subset D, E \in \mathfrak{E} \}$.
\end{definition}

\begin{proposition} \label{prop:uniform_base_symmetric}
The symmetric surroundings form a uniform base.
\end{proposition}
\begin{proof}
We first show for a uniformity $\mathfrak{D}$ that $D \in \mathfrak{D}$ implies $D^{-1} \in \mathfrak{D}$
; From condition (\ref{uniformity_d}), we have $E^{-1} \subset D$ for some $E \in \mathfrak{D}$ so $E \subset D^{-1}$. Hence, $D^{-1} \in \mathfrak{D}$ from condition (\ref{uniformity_e}). 

Next, $D \in \mathfrak{D}$ implies from condition (\ref{uniformity_b}) that $D \cap D^{-1} \in \mathfrak{D}$, which is symmetric. 
\end{proof}

\begin{proposition} \label{prop:uniform_base_conditions}
Suppose $\mathfrak{E}$ is a collection of subsets of $X \times X$. Then $\mathfrak{E}$ is a basis for some uniformity on $X$ if and only if $\mathfrak{E}$ satisfies conditions (\ref{uniformity_a}), (\ref{uniformity_c}), and (\ref{uniformity_d}) as well as the below modified version of (\ref{uniformity_b}):
\begin{align}
    A_1, A_2 \in \mathfrak{E} \Longrightarrow B \subset A_1 \cap A_2 \text{ for some } B \in \mathfrak{E}. \label{uniformity_b_mod}
\end{align}
\end{proposition}
\begin{proof}
We first show that $\mathfrak{D} = \{D \subset X \times X \mid E \subset D, E \in \mathfrak{E} \}$ is a uniformity on $X$ when $\mathfrak{E}$ satisfies all stated conditions. In what follows, let $A_1, A_2 \in \mathfrak{D}$ be arbitrary. Then, there exists $B_1, B_2 \in \mathfrak{E}$ such that $B_1 \subset A_1$ and $B_2 \subset A_2$.\vspace*{0.1cm}\\
\ref{uniformity_a}, \ref{uniformity_c},  \ref{uniformity_d}.\ \ Obvious by the definition of uniform base.\vspace*{0.1cm}\\
\ref{uniformity_b}.\ \ By (\ref{uniformity_b_mod}), there is a $B_3 \in \mathfrak{E}$ such that $B_3 \subset B_1 \cap B_2 \subset A_1 \cap A_2$. Since $B_3 \in \mathfrak{E}$ and $B_3 \subset A_1 \cap A_2$, it follows that $A_1 \cap A_2 \in \mathfrak{D}$.\vspace*{0.1cm}\\
\ref{uniformity_e}.\ \  Suppose $A_3$ satisfies $A_1 \subset A_3 \subset X \times X$. Now, $B_1 \subset A_1$ and $B_1 \in \mathfrak{E}$ imply $A_3 \in \mathfrak{D}$. \vspace*{0.1cm}

Next, we show that each condition is necessary for $\mathfrak{D}$ to be a uniformity. \vspace*{0.1cm}\\
\ref{uniformity_a}.\ \  Obvious.\vspace*{0.1cm}\\
(\ref{uniformity_b_mod})\ \   Choose $B_1, B_2 \in \mathfrak{E}$ such that there is no $E \in \mathfrak{E}$ such that $E \subset B_1 \cap B_2$. Each $A \in \mathfrak{D}$ contains some $E \in \mathfrak{E}$ which is not contained by $B_1 \cap B_2$, so $A \not \subset B_1 \cap B_2$.\vspace*{0.1cm}\\
\ref{uniformity_c}.\ \  Since $B \subset A$ implies $B \circ B \subset A \circ A$, if there is no $E \in \mathfrak{E}$ such that $E \circ E \subset B$ for some $B \in \mathfrak{E}$, then there is no $D \in \mathfrak{D}$ such that $D \circ D \subset B$. \vspace*{0.1cm}\\
\ref{uniformity_d}.\ \   Since $B \subset A$ implies $B^{-1} \subset A^{-1}$, if there is no $E \in \mathfrak{E}$ such that $E^{-1} \subset B$ for some $B \in \mathfrak{E}$, then there is no $D \in \mathfrak{D}$ such that $D^{-1} \subset B$.
\end{proof}

The notion of a uniformity was used to replace the need for a metric, 
so it is unsurprising that a metric generates a uniformity.

\begin{proposition} \label{prop:metric_uniformity}
Given a metric $d$ on $X$, the following sets form a base for a uniformity on $X$:
    \begin{align}
        D_{\epsilon} \doteq D_{X, d, \epsilon} \doteq \left\{  (x,y) \in X \times X \mid d(x,y) < \epsilon \right\} && \epsilon > 0.
    \end{align}
\end{proposition}
\begin{proof}
We show the collection of sets does indeed form a basis for some uniformity:\vspace*{0.1cm}\\
\ref{uniformity_a}., (\ref{uniformity_b_mod}) and \ref{uniformity_d}. are obvious.\vspace*{0.1cm}\\
\ref{uniformity_c}.\ \  Suppose $(a,b) \in D_{\frac\epsilon2} \circ D_{\frac\epsilon2}$, then there is a $z^* \in X$ such that $d(a,z^*) < \frac\epsilon2$ and $d(z^*,b) < \frac\epsilon2$. By the triangle inequality, we have $d(a,b) \leq d(a,z^*) + d(z^*, b) < \epsilon$, so $D_{\frac\epsilon2} \circ D_{\frac\epsilon2} \subset D_{\epsilon}$.
\end{proof}

\begin{definition}[Metric Uniformity] \label{def:metric_uniformity}
A uniformity generated according to Proposition \ref{prop:metric_uniformity} is referred to as the \textit{metric uniformity} generated by $d$ on $X$ and is denoted as $\mathcal{U}_d(X)$.
\end{definition}

\begin{definition}[Standard Uniformity on Real Numbers] \label{def:standard_uniformity}
 The \textit{standard uniformity on $\mathbb{R}$} is the metric uniformity generated from the standard metric. 
\end{definition}

Different metrics may or may not generate different metric uniformities. 
For example, 
\begin{align}
    d_1(x,y) \doteq \left| x - y \right| && d_2(x,y) \doteq \left| \arctan x - \arctan y \right|
\end{align}
both generate the standard topology, but different uniformities. 

The following simple result is useful when working with restrictions of metrics.

\begin{proposition} \label{lem:restricted_metric_generates_subspace_uniformity2}
Suppose $X$ is a uniform space with metric uniformity generated from the metric $d_X$. Let $A \subset X$ and define the metric $d_A$ on $A$ as $d_A(x,y) \mapsto d_X(x,y)$ for each $ x,y \in A$.
Then the metric uniformity on $A$ generated by $d_A$ is the subspace uniformity on $A$ inherited from $X$ (i.e.\,$\mathcal{U}_{d_A}(A) = \mathcal{U}_X(A)$).
\end{proposition}
\begin{proof}
$\{D_{X, d_X, \epsilon} : \epsilon > 0 \}$ is a uniform base for 
$\mathcal{U}_{d_X}(X)$. 
It follows by the definition of a uniform base that $\{ (A \times A) \cap D_{X, d_X, \epsilon} : \epsilon > 0 \}$ is a uniform base for $\mathcal{U}_X(A)$. However, $(A \times A) \cap D_{X, d_X, \epsilon} = D_{A, d_A, \epsilon}$, and $\{D_{A, d_A, \epsilon} : \epsilon > 0 \}$ is a uniform base for $\mathcal{U}_{d_A}(A)$ by definition. So $\mathcal{U}_{d_A}(A)$ and  $\mathcal{U}_X(A)$ have equivalent uniform bases. Hence, the uniformities are the same.
\end{proof}

Every uniformity defines a topology we call the uniform topology. 

\begin{proposition} \label{prop:uniform_topology}
Suppose $\mathfrak{D}$ is a uniformity on $X$. Then, for each $x \in X$, the following collection of sets:
\begin{align}
    \mathcal{N}_x \doteq \{D[x] : D \in \mathfrak{D} \},\quad\text{where } D[x] \doteq \{y \in X \mid (x,y) \in D\},
\end{align}
forms a neighborhood base at $x$ and defines a topology on $X$. 
The same topology is formed if a uniform base $\mathfrak{E}$ is used in place of $\mathfrak{D}$.
\end{proposition}
\begin{proof}
First, we show, for any uniform base $\mathfrak{E}$, $\mathcal{N}_x^{\mathfrak{E}} \doteq \{E[x] : E \in \mathfrak{E} \} $ 
forms a neighborhood base at $x$. 
For any $E \in \mathfrak{E}$, $\Delta(X) \subset E$, so $x \in E[x]$. 
Next, for $E_1, E_2 \in \mathfrak{E}$, we choose $E_3 \in \mathfrak{E}$ such that $E_3 \subset E_1 \cap E_2$, and observe that $E_3[x] \subset (E_1 \cap E_2)[x] = E_1[x] \cap E_2[x]$. 
Finally, $A \in \mathfrak{E}$ implies a $B \in \mathfrak{E}$ such that $B \circ B \subset A$; therefore, $y \in B[x]$ implies $B[y] \subset A[x]$. 

Let $\mathcal{T}_{\mathfrak{E}}$, $\mathcal{T}_{\mathfrak{D}}$ be the topologies generated by $\mathcal{N}_x^{\mathfrak{E}}$, $\mathcal{N}_x^{\mathfrak{D}}$, respectively. 
$\mathcal{N}_x^{\mathfrak{E}} \subset \mathcal{N}_x^{\mathfrak{D}}$, so $\mathcal{T}_{\mathfrak{E}} \subset \mathcal{T}_{\mathfrak{D}}$. 
We show  $\mathcal{T}_{\mathfrak{E}} = \mathcal{T}_{\mathfrak{D}}$. 
Let $U \in \mathcal{T}_{\mathfrak{D}}$ and $x \in U$. Then, there is $V_x \in \mathcal{N}_x^{\mathfrak{D}}$ such that $V_x \subset U$. However, $V_x \in \mathcal{N}_x^{\mathfrak{D}}$ implies a $D \in \mathfrak{D}$ such that $V_x = D[x]$. Since $D \in \mathfrak{D}$ there is $E \in \mathfrak{E}$ such that $E \subset D$. Then, $E[x] \in \mathcal{N}_x^{\mathfrak{E}}$ satisfies $E[x] \subset V_x \subset U$, so $U \in \mathcal{T}_{\mathfrak{E}} = \mathcal{T}_{\mathfrak{D}}$.
\end{proof}

The \textit{uniform topology} is the one generated from a uniformity according to Proposition \ref{prop:uniform_topology}. 
A topological space is \textit{uniformizable} if its topology can be generated by some uniformity. 
We consider a uniform space to be a topological space with its uniform topology. 
Conveniently, the uniform topology of a uniform subspace and metric uniformity correspond with the topological subspace and metric topology, respectively.

\begin{proposition} \label{prop:uniform_metric_generates_metric_topology}
The uniform topology generated by a metric uniformity is the metric topology.
\end{proposition}
\begin{proof}
Suppose $(X,d)$ is a metric space with uniform base $\mathfrak{E} = \{ \{(x,y) : d(x,y) < \epsilon \} : \epsilon > 0 \}$. 
$\mathcal{N}_x = \{D[x] : D \in \mathfrak{E}\}$, where $D[x] = \{y : d(x,y) < \epsilon \}$, is a neighborhood base at $x$
by Proposition \ref{prop:uniform_topology}. It then follows that 
\(\displaystyle 
    \mathcal{B}_{d}(X) = \bigcup_{x \in X} \mathcal{N}_x.
\)
\end{proof}

\begin{proposition}
The uniform topology generated by the relative uniformity is the subspace topology.
\end{proposition}
\begin{proof}
If $\mathcal{N}_x$ is a neighborhood base and $A \subset X$, then $\{ U \cap A : U \in \mathcal{N}_x \}$ is a neighborhood base for the subspace. One has
\begin{align}
    \left\{ E[x] : E \in \mathcal{U}_X(A) \right\} 
    & = \left\{ (D \cap (A \times A))[x] : D \in \mathcal{U}(X) \right\} 
= \left\{ D[x] \cap A : D \in \mathcal{U}(X) \right\}.
\end{align}
\end{proof}

Each Tychonofff space has at least one uniformity that is compatible with its topology.

\begin{proposition} \label{thm:uniformizable_iff_completely_regular}
A topological space is uniformizable if and only if it is completely regular.
\end{proposition}
\begin{proof}
Proved by \cite{willard} Theorem 38.2.
\end{proof}

Uniformly continuous functions are defined in terms of uniformities.

\begin{definition}[Uniformly Continuous] \label{def:uniformly_continuous}
 Let $X$ and $Y$ be uniform spaces. A function $f: X \to Y$ is \textit{uniformly continuous} if and only if for each $E \in \mathcal{U}(Y)$, there is some $D \in \mathcal{U}(X)$ such that $(x,y) \in D \Rightarrow (f(x), f(y)) \in E$. 
\end{definition}

$C_U((X , \mathfrak{D}) ; (Y , \mathfrak{E}))$ denote the collection of uniformly continuous functions from uniform space $(X, \mathfrak{D})$ to uniform $(Y, \mathfrak{E})$, though we just use $C_U(X;Y)$ when there is no confusion over the uniformities. $C_U(X , \mathfrak{D})$ (or $C_U(X)$) is used if $Y=\mathbb R$ with standard uniformity.

\begin{proposition} \label{prop:uniformly_continuous_is_continuous}
Every uniformly continuous function is continuous with respect to the uniform topologies.
\end{proposition}
\begin{proof}
Let $X$, $Y$ be uniform spaces with uniform topologies $\mathcal{T}^U_X$, $\mathcal{T}^U_Y$; $f \colon X \to Y$ be uniformly continuous; and $O \in \mathcal{T}^U_Y$. We show $f^{-1}(O) \in \mathcal{T}^U_X$. 

Suppose $a \in f^{-1}(O)$. Then, Proposition \ref{prop:uniform_topology} says 
$\mathcal{M}_a = \{D[a] : D \in \mathcal{U}(X) \} $, $\mathcal{N}_{f(a)} = \{E[f(a)] : E \in \mathcal{U}(Y) \} $ are neighborhood bases at $a, f(a)$. 
Since $f(a) \in O$ and $O \in \mathcal{T}^U_Y$, there is a $B \in \mathcal{U}(Y)$ such that $B[f(a)] \in \mathcal{N}_{f(a)}$ and $B[f(a)] \subset O$. 
By the uniform continuity of $f$ there is an $A \in \mathcal{U}(X)$ such that $(x,y) \in A$ implies $(f(x), f(y)) \in B$, so $f(A[a]) \subset B[f(a)]$. We have then shown $A[a] \subset f^{-1}(O)$ and clearly $A[a] \in \mathcal{M}_a$, so $f^{-1}(O) \in \mathcal{T}^U_X$.
\end{proof}

\begin{proposition} \label{lem:uniformly_continuous_restricted_is_uniformly_continuous}
Let $X$ and $Y$ be uniform spaces and suppose $f \colon X \to Y$ is uniformly continuous. If $A$ is a uniform subspace of $X$, then $f\vert _{A}$ is uniformly continuous.
\end{proposition}
\begin{proof}
Recall $\mathcal{U}_X(A) \doteq \left\{ (A \times A) \cap D \mid D \in \mathcal{U}(X) \right\}$ is the subspace uniformity. 
Let $E \in \mathcal{U}(Y)$. By the uniform continuity of $f$, there is a $D \in \mathcal{U}(X)$ such that $(x,y) \in D$ implies  $(f(x), f(y)) \in E$. So, $(x,y) \in (A \times A) \cap D$ implies $(f(x), f(y)) \in E$ and $f\vert _{A}$ is uniformly continuous.
\end{proof}

Domain spaces with finer topologies emit richer classes of continuous functions.
Uniform domain spaces with larger uniformities emit richer
classes of uniformly continuous functions. 

\begin{proposition} \label{prop:finer_uniformities_more_cont_funcs}
Suppose $Y$ is a uniform space; $\mathfrak{D}_1$ and $\mathfrak{D}_2$ are uniformities on $X$; and $\mathfrak{D}_1 \subset \mathfrak{D}_2$. Then $C_U((X, \mathfrak{D}_1); Y) \subset C_U((X, \mathfrak{D}_2); Y)$.
\end{proposition}
\begin{proof}
Follows directly from Definition \ref{def:uniformly_continuous}.
\end{proof}

\begin{proposition} \label{thm:compact_hausdorff_uniformity}
Let $X$ be a compact Hausdorff space. Then,
\begin{enumerate}
    \item $X$ has only one uniformity compatible with its topology, and
    \item every continuous function is uniformly continuous.
\end{enumerate}
\end{proposition}
\begin{proof}
See \cite{willard} Theorems 36.18 and 36.19 and Corollary 36.20.
\end{proof}

Cauchy nets generalize Cauchy sequences for metric spaces
to uniform spaces.

\begin{definition}[Cauchy nets]
Let $X$ be a uniform space. A net $(x_{\lambda})_{\lambda \in \Lambda}$ in $X$ is \textit{Cauchy} if and only if for each $D \in \mathcal{U}(X)$, there is some $\lambda_0 \in \Lambda$ such that $(x_{\lambda_1}, x_{\lambda_2}) \in D$ whenever $\lambda_1, \lambda_2 \geq \lambda_0$.
\end{definition}

When the topology in question is the uniform topology, we have the following result. 

\begin{proposition}
Every convergent net is Cauchy.
\end{proposition}
\begin{proof}
See \cite{willard} Theorem 39.2.
\end{proof}

Just as continuous functions map convergent nets to convergent nets, we have uniformly continuous functions map Cauchy nets to Cauchy nets.

\begin{proposition} \label{prop:uniform_continuous_preserves_cauchy}
Suppose $f \colon X \to Y$ is uniformly continuous and $(x_{\lambda})_{\lambda \in \Lambda}$ is a Cauchy net in $X$. Then $(f(x_{\lambda}))_{\lambda \in \Lambda}$ is a Cauchy net in $Y$.
\end{proposition}
\begin{proof}
Take $E \in \mathcal{U}(Y)$. By uniform continuity, there is a $D \in \mathcal{U}(X)$ such that $(x,y) \in D \Rightarrow (f(x), f(y)) \in E$. Since $(x_{\lambda})_{\lambda \in \Lambda}$ is Cauchy, there is a $\lambda_0 \in \Lambda$ such that $(x_{\lambda_1}, x_{\lambda_2}) \in D$ and hence $(f(x_{\lambda_1}), f(x_{\lambda_2})) \in E$ for $\lambda_1, \lambda_2 \geq \lambda_0$. Thus, $(f(x_{\lambda}))_{\lambda \in \Lambda}$ is Cauchy.
\end{proof}

We now consider extending a uniformly continuous function on a dense subspace to its closure. 
Such an extension exists when the codomain is a \textit{complete} uniform space.

\begin{definition}[Complete uniform space]
A uniform space is called \textit{complete} if every Cauchy net converges.
\end{definition}

\begin{theorem} \label{thm:uniformly_continuous_extension}
Let $A \subset X$ be a uniform subspace, $Y$ be a complete uniform space, and $f \colon A \to Y$ be uniformly continuous. For each $p \in \overline{A}$ choose a net $(x^p_{\lambda})$ in $A$ such that $x^p_{\lambda} \to p$ and define the following function
\begin{align}
    g(p) \doteq \lim_{\lambda} f(x^p_{\lambda}).
\end{align}
Then, $g$ extends $f$ to $\overline{A}$ and is uniformly continuous.
\end{theorem}
\begin{proof}
See \cite{willard} Theorem 39.10.
\end{proof}

The next result is important in our universal 
approximation.
In what follows, let $C_U(X)$ denote the set of real uniformly continuous functions on the uniform space $X$. 

\begin{proposition} \label{lem:continuous_on_compact_restricted_is_uniformly_continuous}
Let $X$ be a compact Hausdorff space and $A$ be a uniform dense subspace of $X$ . Then $C(X)\vert _{A} = C_U(A)$. 
\end{proposition}
\begin{proof}
By Theorem \ref{thm:compact_hausdorff_uniformity} there is a unique uniformity on $X$ compatible with its topology and $C_U(X) = C(X)$. 
By Proposition \ref{lem:uniformly_continuous_restricted_is_uniformly_continuous} $C(X)\vert _{A} = C_U(X)\vert _{A} \subset C_U(A)$. 
By Theorem \ref{thm:uniformly_continuous_extension} and the completeness of $\mathbb{R}$ $C_U(A) \subset C_U(X)\vert _{A} = C(X)\vert _{A}$.
\end{proof}
$X$ is compatible with one uniformity but the topology of $A$ may be compatible with more.

\begin{example} \label{ex:confusing_uniformity}
We use ${\lim\limits_{x \to \pm \infty} \arctan x = \pm \frac{\pi}{2}}$ to create 
the metric for $\overline{\mathbb{R}} \doteq \mathbb{R} \cup \{- \infty, \infty \}$
\begin{align}
    \overline{d}_2(x,y) = \left| \lim_{p \to  x} \arctan p - \lim_{q \to y} \arctan q \right|.
\end{align}
$\mathbb{R}$ is a dense subspace of $\overline{\mathbb{R}}$ so Proposition \ref{lem:continuous_on_compact_restricted_is_uniformly_continuous} applies. 
$\mathbb{R}$ inherits its uniformity from $\overline{\mathbb{R}}$,  but it is {not} the standard uniformity as defined in Definition \ref{def:standard_uniformity}
so we should add the metric to the notation as such: $C_U(\mathbb{R}, \mathcal{U}_{\overline{d}_2}(\mathbb{R}))$. 
It turns out $C_U(\mathbb{R}, \mathcal{U}_{\overline{d}_2}(\mathbb{R}))$ is a strict subset of $C_U(\mathbb{R}, \mathcal{U}_{d_1}(\mathbb{R}))$ ($\mathbb{R}$ with standard uniformity and $d_1$ is defined earlier).
\end{example}

%
Generally, homeomorphisms do not preserve uniform continuity but we do have the
following result.

\begin{proposition} \label{lem:equivalent_uniformly_continuous_functions}
Suppose $X$ and $Y$ are compact uniform spaces; $A \subset X$ and $B \subset Y$ are dense uniform subspaces in $X$ and $Y$, respectively; and $h \colon X \to Y$ is a homeomorphism such that $h(A) = B$. Then $C_U(A, \mathcal{U}_X(A)) = \{f \circ h : f \in C_U(B, \mathcal{U}_Y(B))\}$.
\end{proposition}
\begin{proof}
From Proposition \ref{lem:continuous_on_compact_restricted_is_uniformly_continuous}, we know that $C_U(A) = C(X)\vert_A$ and $C_U(B) = C(Y)\vert_B$.
Also, $C(X) \supset \{f \circ h : f \in C(Y)\}$, thus we need only show $C(X) \subset \{f \circ h : f \in C(Y)\}$.
Further, by symmetry we must have $C(Y) \supset \{g \circ h^{-1} : g \in C(X)\}$. It then follows that
\begin{align}
    C(X) \supset& \{f \circ h : f \in C(Y)\}  \\\nonumber
    \supset& \{f \circ h : f \in \{g \circ h^{-1} : g \in C(X)\}\} \\\nonumber
    =& \{g \circ h^{-1} \circ h : g \in C(X)\} \\\nonumber
    =& C(X)
\end{align}
\end{proof}
Compactness is important criteria as it implies we are working with a particular uniformity. 
Let $X$ be a topological space; $\mathcal{M} \subset C_B(X)$ s.p. and s.s.p. on $X$; and $S$ be the unique compactification of $X$ described in Proposition \ref{lem:spp_equiv_compact} statement 2. As $S$ is compact, there is a unique uniformity compatible with the topology on $S$ by Proposition \ref{lem:continuous_on_compact_restricted_is_uniformly_continuous}, which we denote as $\mathfrak{S}_{\mathcal{M}}(S)$ and similarly the inherited uniform subspace on $X$ from $S$ is $\mathfrak{S}_{\mathcal{M}}(X)$. However, typically $X$ is the intended focus of attention, so we will often exclude $X$ from the notation in order to keep things tidy. So will use $\mathfrak{S}_{\mathcal{M}}$ when there is not risk of confusion.

\begin{proposition} \label{lem:uniformly_continuous_ssp_sp}
Suppose $X$ is a topological space, and let $\mathcal{M} \subset C_B(X)$ separate and strongly separate points on $X$. Then $C_U(X,\mathfrak{S}_{\mathcal{M}})$ separates and strongly separates points on $X$ and $\mathcal{M} \subset C_U(X,\mathfrak{S}_{\mathcal{M}})$.
\end{proposition}
\begin{proof}
By Proposition \ref{lem:spp_equiv_compact}, $\bigotimes \mathcal{M}$ extends to a homeomorphism $\widehat{h}_{\mathcal{M}} \colon S \to \mathfrak{cl}[\bigotimes \mathcal{M}(X)]$, where $\mathfrak{cl}[\cdot]$ denotes closure in $\mathbb{R}^{\mathcal{M}}$, and $S$ is compact.

For each $g \in \mathcal{M}$, define $\widehat{g} \doteq \pi_g \circ \widehat{h}_{\mathcal{M}}$ ($\pi$ is the projection function). Then $\widehat{g}$ is a continuous extension of $g$ to the compact set $S$. Since $X$ is a dense uniform subspace of the compact Hausdorff space $S$, we have by Proposition \ref{lem:continuous_on_compact_restricted_is_uniformly_continuous} that $C_U(X,\mathfrak{S}_{\mathcal{M}}) = C(S)\vert_X \ni \widehat{g}\vert_X = g$. This holds for all $g \in \mathcal{M}$, hence we have $\mathcal{M} \subset C_U(X,\mathfrak{S}_{\mathcal{M}})$. It follows from Proposition \ref{lem:ssp_properties} that $C_U(X,\mathfrak{S}_{\mathcal{M}})$ separates and strongly separates points on $X$. 
\end{proof}
\begin{remark}
If $\mathcal{M} = C_B(X)$, then $C_U(X) = C_B(X)$ and the compactified space $S$ is the Stone-Cech compactification.
\end{remark}

\subsection{Spaces of Measures}

We will introduce two topologies on the positive-finite measures (that induce subspace topologies on the probability measures): (1) the \textit{weak topology} and (2) the  \textit{topology of weak convergence}. 
The topology of weak convergence is the sequential topology generated from the weak topology, so they share the same convergent sequences. 
Lastly, we present some results indicating when a collection of functions s.s.p. on these spaces which will be of particular importance for our universal approximation theorems.

Both topologies use the induced functions on the Borel measure:
The mapping $f^*(\mu) = \int_E f \, d\mu$ for integrable $f$. 
Then, $\mathcal{D}^*$ is defined as $\{f^* : f \in \mathcal{D} \}$. 

\begin{definition}[Weak Topology of Finite Measures] \label{def:weak_topology_measures}
Let $(E, \mathcal{T})$ be a topological space. Then, the \textit{weak topology} on $\mathcal{M}^+(E)$ is defined as $\mathcal{O}_{C_B(E)^*}(\mathcal{M}^+(E))$ and is denoted $\mathcal{T}^W$.
\end{definition}
\begin{remark}
By definition, $\mathcal{T}^{W}$ is the coarsest topology on $\mathcal{M}^+(E)$ such that $C_B(E)^* \subset C(\mathcal{M}^+(E))$.
\end{remark}
Before getting to the second topology, we first define a common convergence criteria for spaces of measures.

\begin{definition}[Weak Convergence, Weak Limit Point] \label{def:weak_convergence}
Let $E$ be a topological space. A sequence of measures $\{ \mu_n \}_{n \in \mathbb{N}}  \subset \mathcal{M}^+(E)$ is said to \textit{converge weakly to} $\mu \in \mathcal{M}^+(E)$, written $\mu_n \Rightarrow \mu$, if $\mu_n \to \mu$ with respect to the weak topology $\mathcal{T}^{W}$. This type of convergence is called \textit{weak convergence}. $\mu$ is a \textit{weak limit point} of $\Gamma \subset \mathcal{M}^+(E)$ if there exists a sequence in $\Gamma$ that converges weakly to $\mu$.
\end{definition}

So a weak limit point is the same as a sequential limit point. 
However, weak convergence can also be understood in terms of $C_B(E)^*$. 

\begin{proposition}
$\mu_n \Rightarrow \mu$ if and only if $f^*(\mu_n) \to  f^*(\mu)$ holds for every $f \in C_B(E)$.
\end{proposition}
\begin{proof}
Assume $\mu_n \Rightarrow \mu$. $C_B(E)^*$ are continuous in the weak topology and any sequence is a net; hence, by proposition \ref{prop:continuity_condition_for_nets}, $\mu_n \to \mu$ implies $f^*(\mu_n) \to  f^*(\mu)$ for each $f \in C_B(E)$.

Conversely, given $\mu \in \mathcal{M}^{+}(E)$ and neighborhood $N_{\mu} \in \mathcal{T}^{W}$, there exist $\mathcal{M}_0 \in \mathcal{R}_0(C_B(E))$ and $\epsilon > 0$ such that 
\begin{align}
    \bigcap_{f \in \mathcal{M}_0} \big\{\nu \in \mathcal{M}^{+}(E) : \left| f^*(\mu) - f^*(\nu) \right| < \epsilon \big\}\subset N_{\mu},
\end{align}
since $C_B(E)^*$ s.s.p. by Definition \ref{def:weak_topology_measures}. 
Since $f^*(\mu_n) \to  f^*(\mu)$ for every $f \in \mathcal{M}_0$, there is some $M_{f, \epsilon} \in \mathbb{N}$ such that  ${f^*(\mu_n) \in (f^*(\mu) - \epsilon, f^*(\mu) + \epsilon)}$ for each $n \geq M_{f, \epsilon}$. Thus, we can choose $M_{\epsilon} = \max_{f \in \mathcal{M}_0} \{M_{f, \epsilon}\} \in \mathbb{N}$ so $\mu_n \in N_{\mu}$ for all $n \geq M_{\epsilon}$ and $\mu_n \Rightarrow \mu$.
\end{proof}

It is often useful to find a subcollection $\mathcal{M} \subset C_B(E)$ that needs only be checked to conclude $\mu_n \Rightarrow \mu$. 

\begin{proposition} \label{prop:blount_ssp_prob_measures}
Suppose that $E$ is a topological space; $\{P_n\} \cup \{P\} \subset \mathcal{P}(E)$;  and {$\displaystyle \mathcal M \subset C_{B}\left(E\right) $} is countable, s.p., s.s.p., is closed under multiplication,  and 
\begin{align}
    \int_E g \, dP_n \rightarrow \int_E g \, dP && \forall g \in \mathcal{M}.
\end{align}
Then $P_n \Rightarrow P$.
\end{proposition}
\begin{proof}
See \cite{blount} Theorem 6. 
\end{proof}

Now we define our second topology based on the weak convergence criteria.

\begin{definition}[Topology of Weak Convergence] \label{def:topology_weak_convergence}
The \textit{topology of weak convergence} on $\mathcal{M}^+(E)$ is defined as the sequentially open sets of $\mathcal{M}^+(E)$ generated from the weak topology\footnote{See Definition \ref{prop:gen_seq_top}}. 
We denote the topology of weak convergence as $\mathcal{T}^{WC}$.
\end{definition}

As the topology of weak convergence is the sequential space generated from the weak topology, the two topologies are equivalent when $E$ is a metrizable space. 
This allows us to view continuity in terms of sequences rather than nets.

\begin{proposition} \label{lem:topologies_measures_facts}
The following are true:
\begin{enumerate}
    \item $\mathcal{T}^W$ and $\mathcal{T}^{WC}$ share the same convergent sequences. 
    \item $\mathcal{T}^W \subset \mathcal{T}^{WC}$.
    \item $f \in C(\mathcal{M}^+(E), \mathcal{T}^{WC})$ if and only if $f$ is sequentially continuous; that is, $\mu_n \Rightarrow \mu$ implies $f(\mu_n) \to f(\mu)$. 
    \item If $E$ is a metrizable space, then $(\mathcal{M}^+(E), \mathcal{T}^{W})$ is metrizable and $\mathcal{T}^{W} = \mathcal{T}^{WC}$.
\end{enumerate}
\end{proposition}
\begin{proof}

By Definition \ref{def:topology_weak_convergence}, $(\mathcal{M}^+(E), \mathcal{T}^{WC})$ is a sequential space. So we have:

(1.) Follows from Proposition \ref{prop:same_convergent_sequences}.

(2.) Follows from Proposition \ref{prop:gen_seq_top}.

(3.) Follows from Proposition \ref{prop:sequential_space_sequential_continuous}.

(4.) The development of the Prohorov metric on $(\mathcal{P}(E), \mathcal{T}^{W})$ is discussed in chapter 3 of \cite{ethier_kurtz} and is extended to $(\mathcal{M}^+(E), \mathcal{T}^{W})$ in Chapter 9 problem 6. The equality of the topologies then follows from Proposition \ref{prop:only_squences_when_first_countable} as metric spaces are sequential spaces.
\end{proof}

The following results about sequences of measures apply to both $\mathcal{T}^{W}$ and $\mathcal{T}^{WC}$ as they share the same convergent sequences.

\begin{proposition}[\cite{dong_kouritzin}; Fact 10.1.19] \label{prop:converge_weakly}
Let $E$ be a topological space. Then, the following statements are true:
\begin{enumerate}
    \item $\mu_1 = \mu_2$ in $\mathcal{M}^{+}(E)$ if and only if $\frac{\mu_1}{\mu_1(E)} = \frac{\mu_2}{\mu_2(E)}$ in $\mathcal{P}(E)$ and $\mu_1(E) = \mu_2(E)$.
    \item $\mu_n \Rightarrow \mu$ if and only if $\lim_{n \to \infty} \mu_n(E) = \mu(E)$ and 
    \begin{align}
        \frac{\mu_n}{\mu_n(E)} \Rightarrow  \frac{\mu}{\mu(E)} \text{ in } \mathcal{P}(E). 
    \end{align}
\end{enumerate}
\end{proposition}
\begin{proof}
(1.) is obvious.  
(2.) Assume $\mu_n \Rightarrow \mu$. $1 \in C_B(E)$ implies 
$
    \lim_{n \to \infty} \mu_n(E) = \mu(E)
$.
Then 
\begin{align}
    \lim_{n \to \infty} f^* \left( \frac{\mu_n}{\mu_n(E)} \right) =  \lim_{n \to \infty}  \frac{f^*(\mu_n)}{\mu_n(E)} = \frac{f^*(\mu)}{\mu(E)} = f^* \left( \frac{\mu}{\mu(E)} \right)
\end{align}
holds for each $f \in C_B(E)$.
Conversely, we have for each $f \in C_B(E)$ that 
\begin{align}
    \lim_{n \to \infty}  f^*(\mu_n) = \lim_{n \to \infty} \mu_n(E) f^* \left( \frac{\mu_n}{\mu_n(E)} \right) = 
    \mu(E) f^* \left( \frac{\mu}{\mu(E)} \right) = f^*(\mu).
\end{align}
\end{proof}

The above result immediately implies a one to one relationship regarding the determining sequential point convergence property, which we state next.

\begin{proposition}
\label{lem:separation_on_probs_implies_separation_on_measures}
Let $E$ be a topological space and $1 \in \mathcal{M} \subset M_B(E)$. Then:
\begin{enumerate}
    \item $\mathcal{M}^*$ separates points on $\mathcal{M}^+(E)$ if and only if $\mathcal{M}^*$ separates points on $\mathcal{P}(E)$.
    \item $\mathcal{M}^*$ determines sequential point convergence on $\mathcal{M}^+(E)$ if and only if $\mathcal{M}^*$ determines sequential point convergence on $\mathcal{P}(E)$.
\end{enumerate}
\end{proposition}
\begin{proof}
Follows from Proposition \ref{prop:converge_weakly}.
\end{proof}

When working with the topology of weak convergence, a useful homeomorphism can be defined that relates positive-finite measures with probability measures.

\begin{proposition} \label{prop:homeomorphism_for_measures}
Equip  $\mathcal{M}^+(E)$ and $\mathcal{P}(E)$ with the topology of weak convergence and consider the function $H \colon \mathcal{M}^+(E) \to (0, \infty) \times \mathcal{P}(E) $ defined as $H(\mu) \mapsto \left( \mu(E), \frac{\mu}{\mu(E)} \right)$. Then $H$ is a homeomorphism with inverse $H^{-1}(c, P) \mapsto c \cdot P$.
\end{proposition}
\begin{proof}
Proposition \ref{prop:converge_weakly} (1) establishes $H$ is a bijection, so it remains to show continuity.
$\mathcal{M}^+(E)$ and $\mathcal{P}(E)$ with the topology of weak convergence are sequential spaces. 
Further, $(0, \infty)$ as a subspace of $\mathbb{R}$ is a locally compact sequential space.
Theorem 4 from \cite{sequential_spaces} says that the product of two sequential spaces is sequential when one of the spaces is locally compact, so $(0, \infty) \times \mathcal{P}(E)$ is a sequential space. Hence, $H$ is a mapping between sequential spaces, so its continuity properties are reduced to checking for sequential continuity, which was established in Proposition \ref{prop:converge_weakly} (2). 
\end{proof}

The above is particularly important for our universal approximation results where we will want bounded functionals. 

Next, we provide a result demonstrating the difficulty in trying to find a collection of functionals that s.s.p. on the topology of weak convergence. 

\begin{proposition}
$C_B(E)^*$ always s.s.p. on $(\mathcal{M}^+(E), \mathcal{T}^{W})$; however, $C_B(E)^*$ s.s.p. on $(\mathcal{M}^+(E), \mathcal{T}^{WC})$ if and only if $\mathcal{T}^{WC} = \mathcal{T}^{W}$.
\end{proposition}
\begin{proof}
Proposition \ref{prop:ssp_equiv_to_finer_topology} says that $\mathcal{M}$ s.s.p. on $(X, \mathcal{T})$ if and only if $\mathcal{T} \subset \mathcal{O}_{\mathcal{M}}(X)$. So the result follows from Definition \ref{def:weak_topology_measures} and Proposition \ref{lem:topologies_measures_facts} (1), where we have $\mathcal{O}_{C_B(E)^*}(\mathcal{M}^+(E)) = \mathcal{T}^{W} \subset \mathcal{T}^{WC}$.
\end{proof}

We focus largerly on metrizable $E$ so 
the weak topology and topology of weak convergence are the same 
and we only need to consider converging sequences rather than nets 
(see Proposition \ref{lem:topologies_measures_facts}). 
There is a nice way of determining the s.s.p. property on measures spaces.

\begin{theorem} \label{thm:blount_ssp_prob_measures}
Suppose that $E$ is a topological space; $g_0 \in C_B((0, \infty))$ s.p. and s.s.p. on $(0, \infty)$;  and {$\displaystyle \mathcal M = \{1\} \cup \{g_i \}_{i=2}^N \subset C_{B}\left(E\right) $} is countable, s.p., s.s.p., and is closed under multiplication.
Further, define $\mathfrak{W}_{g_0}[\mathcal{M}] \doteq \{ \mu \mapsto g^*( \frac{\mu}{\mu(E)}) : g \in \mathcal{M}, g \neq 1 \} \cup \{\mu \mapsto g_0(\mu(E)) \}$. Then
\begin{enumerate}
    \item $\mathcal{M}^+(E)$ is metrizable and $\mathcal{T}^{WC} =  \mathcal{T}^{W}$,
    \item $\mathfrak{W}_{g_0}[\mathcal{M}] \subset C_B(\mathcal{M}^+(E))$
    \item $\mathfrak{W}_{g_0}[\mathcal{M}]$ determines sequential point convergence on $\mathcal{M}^+(E)$,
    \item $\mathfrak{W}_{g_0}[\mathcal{M}]$ s.s.p and s.p. on $\mathcal{M}^+(E)$,
    \item $\bigotimes \mathfrak{W}_{g_0}[\mathcal{M}] \colon \mathcal{M}^+(E) \to \bigotimes \mathfrak{W}_{g_0}[\mathcal{M}](\mathcal{M}^+(E))$ is a homeomorphism.
\end{enumerate}
\end{theorem}

\begin{proof}
(1.) By Proposition \ref{lem:countable_sep_spp_has_this_metric}, $E$ is metrizable, so $\mathcal{M}^+(E)$ is metrizable (hence, Hausdorff) and $\mathcal{T}^{WC} =  \mathcal{T}^{W}$ by Proposition \ref{lem:topologies_measures_facts}.

(2.) Boundedness of $\mu \mapsto g^*( \frac{\mu}{\mu(E)})$ and $\mu \mapsto g_0(\mu(E))$ comes from the $g_i$ boundedness. $\mathfrak{W}_{g_0}[\mathcal{M}] \subset C_B(E)^* \subset C(\mathcal{M}^+(E))$ since $g_0 $ and $\mathcal M$ are bounded and continuous on $E$.

(3.) $\{g_i^* \}_{i=2}^N$ determines sequential point convergence on $\mathcal{P}(E)$ by Proposition \ref{prop:blount_ssp_prob_measures}. Also, $g_0 \in C_B((0, \infty))$ determines sequential point convergence on $(0, \infty)$ as it s.p. and s.s.p. so 
\begin{align}
    \{ g_0 \circ \pi_1 \} \cup \{ g_i^* \circ \pi_2 \}_{i=2}^N 
\end{align}
determines sequential point convergence on $(0, \infty) \times \mathcal{P}(E)$.
The rest follows from the homeomorphism $H \colon \mathcal{M}^+(E) \to (0, \infty) \times \mathcal{P}(E)$ of Proposition \ref{prop:homeomorphism_for_measures}.

(4.) S.s.p. is implied by (1.), (3.), Proposition \ref{prop:ssp_homeomorphism_determine_point_convergence} ($4 \to 2$), and that $\mathfrak{W}_{g_0}[\mathcal{M}]$ is countable. (1.) implies $\mathcal{M}^+(E)$ is Hausdorff and combined with Proposition \ref{lem:spp_implies_sp_on_hausdorff}, implies $\mathfrak{W}[\mathcal{M}]$ s.p..

(5.) Follows from (2.), (4.), and Proposition \ref{prop:ssp_homeomorphism_determine_point_convergence} ($2 \to 1$).
\end{proof}

We finish this section with a result about uniformly continuous functions.

\begin{proposition} \label{thm:uniformly_continuous_ssp_of_prob_measures}
Suppose $E$ is a topological space; $g_0 \in C_B((0, \infty))$ s.p. and s.s.p. on $(0, \infty)$; and let $\mathcal{M} \subset C_B(E)$ be countable, s.p. and s.s.p. on $E$. Then $\mathfrak{W}_{g_0}[C_U(E,\mathfrak{S}_{\mathcal{M}})]$ s.p. and s.s.p. on $\mathcal{M}^+(E)$.
\end{proposition}
\begin{proof}
By Proposition \ref{lem:uniformly_continuous_ssp_sp} $C_U(E,\mathfrak{S}_{\mathcal{M}}(E))$ s.p. and s.s.p. It is also closed under multiplication, while $\mathcal{M} \subset C_U(E,\mathfrak{S}_{\mathcal{M}})$ is of size $N \in \mathbb{N} \cup \{\infty\}$.
Let $\mathcal{M} = \{g_i \}_{i=1}^N$ and
\begin{align}
    \mathcal{N}_i = \left\{ \prod_{g \in \mathcal{C}_0} g : \mathcal{C}_0 \in \mathcal{R}_0 \left(\{g_j \}_{j=1}^i \right)   \right\},
\end{align}
which is finite for each $i$, so $\mathcal{N} = \bigcup_{i=1}^N \mathcal{N}_i$ is countable, closed under multiplication, and contains $\mathcal{M}$. 
Hence, $\mathcal{N} \subset C_B(E)$ is countable, closed under multiplication, s.p., and s.s.p. on $E$, so $\mathfrak{W}_{g_0}[\mathcal{N}]$ s.s.p. and s.p. on $\mathcal{M}^+(E)$ by Proposition \ref{thm:blount_ssp_prob_measures}.

Further, $\mathcal{N} \subset C_U(E,\mathfrak{S}_{\mathcal{M}})$ since $C_U(E,\mathfrak{S}_{\mathcal{M}})$ is closed under multiplication and contains $\mathcal{M}$. The result then follows from $\mathfrak{W}_{g_0}[\mathcal{N}] \subset \mathfrak{W}_{g_0}[C_U(E,\mathfrak{S}_{\mathcal{M}})]$.
\end{proof}

\section{Universal Approximation Results} \label{sec:hausdorff}

We now provide universal approximation results for Tychonoff spaces and spaces of measures. 
First, a uniform dense result on compact Hausdorff spaces is established by 
a version of the Stone-Weierstrass Theorem from \cite{rudin_FA}. 

\begin{theorem}[Stone-Weierstrass]   \label{thm:stone}
Let $X$ be a compact Hausdorff space and let $C(X)$ be the set of real continuous functions on $X$ equipped with the sup metric. Suppose that:
\begin{enumerate}
    \item $A$ is a closed subalgebra of $C(X)$,
    \item $A$ separates points on $X$,
    \item $A$ vanishes nowhere on $X$ (i.e., at every $p \in X$, $f(p) \neq 0$ for some $f \in A$).
\end{enumerate}
Then $A = C(X)$.
\end{theorem}

In fact, the algebra s.s.p. on $X$ as we saw in Proposition \ref{lem:compact_hausdorff_spp_equiv_sp} s.p. implies s.s.p. in a compact Hausdorff setting.

Given a subset of continuous functions $\mathcal{M} \subset C(X)$ closed under addition, we can construct an algebra of continuous functions using the exponential function.

\begin{lemma}
Suppose $X$ is a topological space and $\mathcal{M} \subset C(X)$ is closed under addition. Then the following collection of functions
\begin{align}
    \Lambda(\mathcal{M}) \doteq \bigg\{ p \mapsto \sum_{i=1}^n c_i e^{g_i(p)} \mid g_i \in \mathcal{M}; c_i \in \mathbb{R}; n \in \mathbb{N}  \bigg\}, \label{eqn:exponentials}
\end{align}
is an algebra and $\Lambda(\mathcal{M}) \subset C(X)$.
\end{lemma}
\begin{proof}
Follows immediately from direct verification.
\end{proof}

The Stone-Weierstrass Theorem establishes when our algebra is uniform dense in $C(X)$.

\begin{lemma}\label{lem:exponentials_uniform_dense}
Let $X$ be a compact Hausdorff space, and let $\mathcal{M} \subset C(X)$ be closed under addition and separate points on $X$. Then $\Lambda(\mathcal{M})$ (defined by \ref{eqn:exponentials}) is uniform dense in $C(X)$.
\end{lemma}
\begin{proof}
$\mathcal{M}$ is closed under addition, so $\Lambda(\mathcal{M})$ is a subalgebra of $C(X)$. 
$e^{g(p)} > 0$ for all $g \in \mathcal{M},p \in X$, so $\Lambda(\mathcal{M})$ vanishes nowhere. 
$\Lambda(\mathcal{M})$ s.p. since 
$\mathcal{M}$ does and the exponential function is injective.
\end{proof}

We now provide our universal approximation on compact Hausdorff spaces. 

\begin{theorem} \label{thm:dense_in_lambda}
Let $X$ be a compact Hausdorff space, and let $\mathcal{M} \subset C(X)$ separate points on $X$. Suppose that, for every $n \in \mathbb{N}$, $\mathcal{F}_n\subset C(\mathbb{R}^n)$ is uniform dense on compacts of $\mathbb{R}^n$. Then the following set is a uniform dense subset of $C(X)$:
\begin{align}
    \mathfrak{N}( \mathcal{M}, \{ \mathcal{F}_n \}_{n=1}^{\infty}) \doteq \big\{ p \mapsto f  (g_1(p),\, \ldots \, , g_n(p)) :  n \in \mathbb{N} ; \; f \in \mathcal{F}_n; \; \{g_i\}_{i=1}^n \in \mathcal{R}_0(\mathcal{M})  \big\}.
\end{align}
\end{theorem}
\begin{proof}
Clearly, $\mathfrak{H}( \mathcal{M}) \subset C(X)$ so it remains to show $\mathfrak{H}( \mathcal{M})$ is uniform dense in $C(X)$. 

We wish to employ Lemma \ref{lem:exponentials_uniform_dense} and the transitive property of dense sets; however, Lemma \ref{lem:exponentials_uniform_dense} assumes $\mathcal{M}$ is closed under addition so we construct:
\begin{align}
    \mathcal{M}^{(+)} \doteq \bigg\{ \sum_{g \in H} g \mid H \subset \mathcal{M}, \; H  \text{ finite} \bigg\}.
\end{align}
$\mathcal{M}^{(+)}$ is closed under addition, inherits the s.p. property and hence $\Lambda(\mathcal{M}^{(+)})$ is dense in $C(X)$ by Lemma \ref{lem:exponentials_uniform_dense}.

We show $\mathfrak{N}( \mathcal{M}, \{ \mathcal{F}_n \}_{n=1}^{\infty})$ is dense in $\Lambda(\mathcal{M}^{(+)})$. 
Each $\lambda \in \Lambda(\mathcal{M}^{(+)})$ takes the form:
\begin{align}
        \lambda &= \sum_{i=1}^m c_i \exp{ \{ h_i \} }  &&  h_i \in \mathcal{M}^{(+)}, \; c_i \in \mathbb{R} \\ \nonumber
                &= \sum_{i=1}^m c_i \exp{ \bigg\{ \sum_{g \in H^{(i)}} g \bigg\} } && H^{(i)} \subset \mathcal{M}, \; |H^{(i)} | < \infty.
\end{align}
Letting $n = \left|\bigcup_{i=1}^m H^{(i)} \right|$, we can rewrite $\lambda$ as a composition of continuous functions $\Gamma:X \to \mathbb{R}^n$, $\Phi:\mathbb{R}^n \to \mathbb{R}^m$, and $\Psi:\mathbb{R}^m \to \mathbb{R}$ defined as
\begin{align}
    \Gamma(x) &\mapsto ( g_i(x) )_{i=1}^n  &&  \\ 
    \Phi((y_i)_{i=1}^n) &\mapsto \bigg( \sum_{i=1}^n y_i I_{[g_i \in H^{(j)}]} \bigg)_{j=1}^m  && \\  
    \Psi( (z_i)_{i=1}^m ) &\mapsto  \sum_{i=1}^m c_i e^{z_i},
\end{align}
so $\lambda = \Psi \circ \Phi \circ \Gamma$.

$X$ is compact and $\Gamma$ is continuous, so $\Gamma(X) \subset \mathbb{R}^n$ is compact. Since $\Psi \circ \Phi \in C(\mathbb{R}^n)$, we have for each $\epsilon > 0$, there exists a function $f \in \mathcal{F}_n$ such that
\begin{align}
    \epsilon &> \sup \left\{\,\left| (\Psi \circ \Phi)(y) - f(y)\right|:y\in \Gamma(X) \,\right\} \\\nonumber
             &= \sup \left\{\,\left| (\Psi \circ \Phi \circ \Gamma)(x) - (f \circ \Gamma)(x)\right|:x\in X \,\right\} \\\nonumber
             &= \sup \left\{\,\left| \lambda(x) - (f \circ \Gamma)(x)\right|:x\in X \,\right\}.
\end{align}
By definition $f \circ \Gamma \in \mathfrak{N}( \mathcal{M}, \{ \mathcal{F}_n \}_{n=1}^{\infty})$, so $\mathfrak{N}( \mathcal{M}, \{ \mathcal{F}_n \}_{n=1}^{\infty})$ is dense in $\Lambda(\mathcal{M}^{(+)})$ and $C(X)$.
\end{proof}


To eventually use
the universal approximation for compact spaces we just developed
we will have to compactify Tychonoff spaces. The following is our main topological universal approximation result, which
was stated in an earlier section.

\begin{theorem}\label{thm:neural_networks_on_tychonoff}
Suppose $X$ is a topological space; $\mathcal{M} \subset C_B(X)$ separate and strongly separate points on $X$; and, for each $n \in \mathbb{N}$, $\mathcal{F}_n$ is uniform dense on compacts of $\mathbb{R}^n$. Then $\mathfrak{N}( \mathcal{M}, \{ \mathcal{F}_n \}_{n=1}^{\infty})$
is a uniform dense subset of $C_U(X,\mathfrak{S}_{\mathcal{M}})$. Additionally, if $\mathcal{M}$ is countable with cardinality $N \in \mathbb{N} \cup \{\infty\}$, then $\mathfrak{S}_{\mathcal{M}}$ is equivalent to the metric uniformity generated by the following metric:
\begin{align}
    d(x,y) \mapsto \sum_{i=1}^{N} 2^{-i} \left( |g_i(x) - g_i(y)| \wedge 1 \right) && \forall x,y \in X.
\end{align}
\end{theorem}
\begin{proof}
By Proposition \ref{lem:spp_equiv_compact}, $\bigotimes \mathcal{M}$ extends to a homeomorphism $\widehat{h}_{\mathcal{M}} \colon S \to \mathfrak{cl}[\bigotimes \mathcal{M}(X)]$, where $\mathfrak{cl}[\cdot]$ denotes closure in $\mathbb{R}^{\mathcal{M}}$ and $S$ is compact.

For each $g \in \mathcal{M}$, define $\widehat{g} \doteq \pi_g \circ \widehat{h}_{\mathcal{M}}$. Then $\widehat{g}$ is a continuous extension of $g$ to the compact set $S$. Define $\widehat{\mathcal{M}} \doteq \{\widehat{g} : g \in \mathcal{M} \}$. We find that $ \bigotimes \widehat{\mathcal{M}}$ is $\widehat{h}_{\mathcal{M}}$, so it is a homeomorphism and by Proposition $\ref{lem:spp_equiv_compact}$ we see that $\widehat{\mathcal{M}}$ separates points (and strongly separates points) on $S$.

Therefore, $\mathfrak{N}( \widehat{\mathcal{M}},\{ \mathcal{F}_n \}_{n=1}^{\infty})$ is a uniform dense subset of $C(S)$ by Theorem \ref{thm:dense_in_lambda} so $\mathfrak{N}( \mathcal{M},\{ \mathcal{F}_n \}_{n=1}^{\infty})$ is a uniform dense subset of $C(S)\vert _{X} = C_U(X,\mathfrak{S}_{\mathcal{M}})$ by Proposition \ref{lem:continuous_on_compact_restricted_is_uniformly_continuous}.

When $\mathcal{M}$ is countable, it follows by Proposition \ref{lem:countable_sep_spp_has_this_metric} that $S$ is metrized by:
\begin{align}
    \widehat{d}(x,y) \mapsto \sum_{i=1}^{N} 2^{-i} \left( |\widehat{g}_i(x) - \widehat{g}_i(y)| \wedge 1 \right) && \forall x,y \in S.
\end{align}
Proposition \ref{thm:compact_hausdorff_uniformity} (1) implies that $\mathfrak{S}_{\mathcal{M}}(S)$ is exactly the metric uniformity $\mathcal{U}(S;\widehat{d})$ as it is unique.
Since $\widehat{g}_i\vert_X = g_i$ for each $i$, the metric $d$ is just
\begin{align}
    d(x,y) \mapsto \widehat{d}(x,y) && \forall x,y \in X.
\end{align}
So by Proposition \ref{lem:restricted_metric_generates_subspace_uniformity2} the subspace uniformity on $X$ inherited from $S$ is the metric uniformity generated by $d$, which is to say that $\mathcal{U}_S(X) = \mathfrak{S}_{\mathcal{M}} = \mathcal{U}_d(X)$.
\end{proof}
\begin{remark}
Proposition \ref{lem:tychonoff_equiv_sep_and_ssp} tells us $X$ has to be a Tychonoff space as there is some collection of continuous functions that separate and strongly separate points on $X$.
\end{remark}

The result demonstrates that we can approximate uniformly continuous functions from a unique uniformity which has an associated metric in the case where we have a countable collection of functions that strongly separate points.

\subsection{Spaces of Measures}

We now apply Theorem \ref{thm:neural_networks_on_tychonoff} to spaces of measures. 
Given topological space $E$; bounded function $g_0$; and collection $\mathcal{M} \subset C_B(E)$, recall functionals on $\mathcal{M}^+(E)$
\begin{align}
    \mathfrak{W}_{g_0}[\mathcal{M}] \doteq \left\{ \mu \mapsto g^*\left( \frac{\mu}{\mu(E)} \right) : g \in \mathcal{M}, g \neq 1 \right\} \cup \{\mu \mapsto g_0(\mu(E)) \}.
\end{align}


\begin{theorem} \label{thm:universal_approximation_of_measures_theoretical}
Suppose $N\in\mathbb N\cup\{\infty\}$; $E$ is a topological space;  {$\displaystyle g_{0} \in C_B(E) $} s.p. and s.s.p. on {$\displaystyle (0, \infty) $}; {$\displaystyle \mathcal M = \{g_i \}_{i=1}^N \subset C_{B}\left(E\right) $} s.p., s.s.p., is countable and closed under multiplication; and {$\displaystyle \mathcal F_{n}\subset C(\mathbb R^{n}) $} is uniform dense on the compacts of {$\displaystyle \mathbb{R}^{n} $} for each {$\displaystyle n\in \mathbb{N} $}. 
Then $\mathfrak{D}_{g_0}(\mathcal{\mathcal{M}}, \{ \mathcal{F}_n \}_{n=1}^{\infty})$ is a uniform dense subset of {$\displaystyle C_U \big(\mathcal{M}^+(E),\mathfrak{S}_{\mathfrak{W}_{g_0}[\mathcal{M}]} \big) $}. Additionally, $\mathfrak{S}_{\mathfrak{W}_{g_0}[\mathcal{M}]}$ is equivalent to the metric uniformity generated by the metric:
\begin{align}
    d(\mu,\nu) \mapsto 
    \left(
    \begin{aligned}
    &|g_0(\mu(E)) - g_0(\nu(E))| \wedge 1 \\ + \sum_{i=2}^N 2^{-i}& \left( \left| g_i^*\left( \frac{\mu}{\mu(E)} \right) - g_i^*\left( \frac{\nu}{\nu(E)} \right) \right| \wedge 1 \right)
    \end{aligned}
    \right)
\end{align}
for each $ \mu,\nu \in \mathcal{M}^+(E).$
\end{theorem}
\begin{proof}
By Proposition \ref{thm:blount_ssp_prob_measures}, $\mathfrak{W}_{g_0}[\mathcal{M}] \subset C_B(\mathcal{M}^+(E))$ is countable, s.s.p., and s.p. on $\mathcal{M}^+(E)$. So the result follows directly from Theorem \ref{thm:neural_networks_on_tychonoff}.
\end{proof}

Proposition \ref{lem:dense_sets_still_sp_or_spp} said that if $\mathcal{M}$ 
s.s.p. or s.p. and $\mathcal{M}_0$ is uniform dense in $\mathcal{M}$, then 
$\mathcal{M}_0$ s.s.p. or s.p. also. 
The next lemma uses this proposition to produce a similar result for spaces of measures. 

\begin{lemma} \label{dense_funcs_for_measures}
Let $E$ be a metrizable topological space and assume $\mathcal{M}, \mathcal{M}_0 \subset C_B(E)$; $\mathcal{M}^*$ s.s.p. and s.p. on $\mathcal{P}(E)$; and $\mathcal{M}_0$ is uniform dense in $\mathcal{M}$. Then $\mathcal{M}_0^*$ s.s.p. and s.p. on $\mathcal{P}(E)$. Further, given $ g_{0} \in C_B((0, \infty)) $ and {$\displaystyle \mathcal F_{n} \subset C(\mathbb R^{n}) $} for each $n \in \mathbb{N}$, we find that $\mathfrak{D}_{g_0}(\mathcal{M}_0, \{ \mathcal{F}_n \}_{n=1}^{\infty})$ is uniform dense in $\mathfrak{D}_{g_0}(\mathcal{\mathcal{M}}, \{ \mathcal{F}_n \}_{n=1}^{\infty})$. 
\end{lemma}
\begin{proof}
Given $f \in \mathcal{M}$ and $\epsilon > 0$, there is a $g \in \mathcal{M}_0$ such that
\begin{align}
    \epsilon > \sup\{|f(p) - g(p)| : p \in E\}.
\end{align}
since $\mathcal{M}_0$ is uniform dense in $\mathcal{M}$.
Therefore, 
\begin{align}
    \left| \int_E f \, dP - \int_E g \, dP \right| &\leq   \int_E |f(p) - g(p)| \, dP < \epsilon
\end{align}
for each $P \in \mathcal{P}(E)$. 
Hence, $\mathcal{M}_0^*$ is uniform dense in $\mathcal{M}^*$, so it s.s.p. and s.p. on $\mathcal{P}(E)$ too by Proposition \ref{lem:dense_sets_still_sp_or_spp}. Picking $g_0$ that s.p., s.s.p., we find $\mathfrak{W}_{g_0}[ \mathcal{M}_0]$ s.s.p., s.p. on $\mathcal{M}^+(E)$.

Now, for $n \in \mathbb N$; $\big\{\widetilde{f}_i \big\}_{i=2}^n \subset \mathfrak{W}_{g_0}[\mathcal{M}]$ and $h \in C(\mathbb R^n)$; $\psi \in \mathfrak{D}_{g_0}(\mathcal{\mathcal{M}}, \{ \mathcal{F}_n \}_{n=1}^{\infty})$ has form:
\begin{align}
    \psi&(\mu) \mapsto h\left(g_0(\mu(E)),\, \widetilde{f}_2(\mu), \ldots, \widetilde{f}_n(\mu)\right). 
\end{align}
But, if $\mathcal{N} = \{\mu \mapsto g_0(\mu(E)) \} \cup \big\{\widetilde{f}_i \big\}_{i=2}^n$, then $\bigotimes \mathcal{N}(\mathcal{M}^+(E)) \subset K$ for some compact set $K \subset \mathbb{R}^n$. 
Hence, $h \vert_K$ is a continuous function on a compact set, so by Proposition \ref{thm:compact_hausdorff_uniformity}, there is a unique uniformity compatible with the topology on $K$ and $h \vert_K$ is uniformly continuous. This topology is the metric (inherited from $\mathbb R^n$) uniformity. Therefore, for any $\epsilon > 0$, there is a $\delta_{\epsilon} > 0$ such that
\begin{align}
    d(p,q) < \delta_{\epsilon} \;\; \text{ implies } \;\; |h(p) - h(q)| < \epsilon && p,q \in K,
\end{align}
where $d$, which can be any metric that generates the topology on $K$; is chosen to be
\begin{align}
    d(p,q) \mapsto \sum_{i=1}^n |\pi_i(p) - \pi_i(q)|.
\end{align}
We found that for each ${f}^*_i$, there is a ${g}^*_i \in \mathfrak{W}_{g_0}[\mathcal{M}_0]$ such that 
$\displaystyle
    \sup\left\{\left| {f}^*_i(\mu) - {g}^*_i(\mu) \right| : \mu \in \mathcal{P}(E) \right\} < \frac{\delta_{\epsilon}}{n}
$.
Letting $\mathcal{N}_0 = \{ \mu \mapsto g_0(\mu(E)) \} \cup \big\{{g}^*_i \big\}_{i=2}^n$, we have
\begin{align}
    d \left(\bigotimes \mathcal{N}(\mu) , \bigotimes \mathcal{N}_0(\mu)  \right) &=
     \sum_{i=2}^n |\widetilde{f}_i(\mu) - \widetilde{g}_i(\mu)|< \delta_{\epsilon}
\end{align}
holds for all $\mu \in \mathcal{P}(E)$. Hence, $\mathfrak{D}_{g_0}(\mathcal{M}_0, \{ \mathcal{F}_n \}_{n=1}^{\infty})$ is uniform dense in $\mathfrak{D}_{g_0}(\mathcal{\mathcal{M}}, \{ \mathcal{F}_n \}_{n=1}^{\infty})$. 
\end{proof}

Notice the "closed under addition constraint" has been removed from 
$\mathcal{M}$, which can facilitate use.
\begin{theorem} \label{thm:universal_approximation_of_measures_built_from_topological_nns}
Suppose $E$ is a topological space;  $g_{0} \in C_B(E) $ s.p. and s.s.p. on $(0, \infty)$; $ \mathcal M =  \{g_i \}_{i=1}^N  $ is countable, s.p. and s.s.p. on $E$; and, for each $n \in \mathbb{N}$, $\mathcal F_{n}, \mathcal{H}_n \subset C(\mathbb R^{n}) $ are uniform dense on the compacts of $\mathbb{R}^{n}$.
Then
$\mathfrak{D}_{g_0}( \mathfrak{N} (\mathcal{M}, \{ \mathcal{F}_n \}_{n=1}^{\infty}), \{ \mathcal{H}_n \}_{n=1}^{\infty})$ is a uniform dense subset of {$\displaystyle C_U \big(\mathcal{M}^+(E),\mathfrak{S}_{\mathfrak{W}_{g_0}[C_U(E,\mathfrak{S}_{\mathcal{M}})]} \big) $}.
\end{theorem}

\begin{proof} 
By Theorem \ref{thm:uniformly_continuous_ssp_of_prob_measures}, $\mathfrak{W}_{g_0}[C_U(E,\mathfrak{S}_{\mathcal{M}})]$ are bounded continuous functions that s.s.p. and s.p. on $\mathcal{M}^+(E)$, so it then follows by Theorem \ref{thm:universal_approximation_of_measures_theoretical} that $\mathfrak{D}_{g_0}( C_U(E,\mathfrak{S}_{\mathcal{M}}), \{ \mathcal{H}_n \}_{n=1}^{\infty})$ is a uniform dense subset of {$\displaystyle C_U \big(\mathcal{M}^+(E),\mathfrak{S}_{\mathfrak{W}_{g_0}[C_U(E,\mathfrak{S}_{\mathcal{M}})]} \big) $}.  Also, Theorem \ref{thm:neural_networks_on_tychonoff} says $\mathfrak{N} (\mathcal{M},\{ \mathcal{F}_n \}_{n=1}^{\infty})$
is a uniform dense subset of $C_U(E,\mathfrak{S}_{\mathcal{M}})$. Hence, the result follows from Lemma \ref{dense_funcs_for_measures} (with $\mathcal{M}_0 = \mathfrak{N} (\mathcal{M},\{ \mathcal{F}_n \}_{n=1}^{\infty})$ and $\mathcal{M} = C_U(E,\mathfrak{S}_{\mathcal{M}})$).
\end{proof}


\bibliography{references}
\bigskip

\end{document}